\documentclass[12pt]{spieman}  
\usepackage{amsmath,amsfonts,amssymb}
\usepackage{graphicx}
\usepackage{setspace}
\usepackage{tocloft}

\usepackage{cite}
\usepackage[utf8]{inputenc}
\usepackage{booktabs}
\usepackage{url}
\usepackage[acronym]{glossaries}
\usepackage{gensymb} 
\newacronymstyle{long-short-br}
{%
  \GlsUseAcrEntryDispStyle{long-short}%
}%
{%
  \GlsUseAcrStyleDefs{long-short}%
}
\setacronymstyle{long-short-br}

\usepackage{xcolor}
\usepackage[caption=false, position=top]{subfig}
\usepackage{balance}
\usepackage{multirow}
\usepackage{xspace} 
\usepackage{arydshln} 
\newcommand{\minus}{\scalebox{0.75}[1.0]{$-$}}

\usepackage[right, modulo]{lineno}

\usepackage{fancyhdr}

\newcommand\REV[1]{{#1}}
\newcommand\MINOR[1]{{#1}}

\title{Convolutional Neural Networks for Automatic Meter Reading}

\author[a,*]{Rayson Laroca}
\author[a]{Victor Barroso}
\author[b]{Matheus A. Diniz}
\author[b]{Gabriel R. Gonçalves}
\author[b]{William~Robson Schwartz}
\author[a]{David Menotti}
\affil[a]{Federal University of Paran\'a, Laboratory of Vision, Robotics and Imaging, Department of Informatics, Av. Coronel Francisco Heráclito dos Santos 100, Curitiba, Brazil, 81530-000}
\affil[b]{Federal University of Minas Gerais, Smart Surveillance Interest Group, Department of Computer Science, Av. Antônio Carlos 6627, Belo Horizonte, Brazil, 31270-010} 

\cftpagenumbersoff{figure}
\cftpagenumbersoff{table} 
\begin{document} 

\sloppy 

\maketitle

\newacronym{amr}{AMR}{Automatic Meter Reading}
\newacronym{ann}{ANN}{Artificial Neural Network}
\newacronym{bflop}{BFLOP}{billion floating-point operations}
\newacronym{cca}{CCA}{Connected Components Analysis}
\newacronym{cnn}{CNN}{Convolutional Neural Network}
\newacronym{copel}{Copel}{Energy Company of Paraná}
\newacronym{crnn}{CRNN}{Convolutional Recurrent Neural Network}
\newacronym{ctc}{CTC}{Connectionist Temporal Classification}
\newacronym{fps}{FPS}{frames per second}
\newacronym{hog}{HOG}{Histogram of Oriented Gradients}
\newacronym{iou}{IoU}{Intersection over Union}
\newacronym{lpr}{LPR}{License Plate Recognition}
\newacronym{lstm}{LSTM}{long short-term memory}
\newacronym{ocr}{OCR}{Optical Character Recognition}
\newacronym{roi}{ROI}{Region of Interest}
\newacronym{fcn}{FCN}{Fully Convolutional Network}
\newacronym{mlp}{MLP}{Multilayer Perceptron}
\newacronym{mser}{MSER}{Maximally Stable Extremal Regions}
\newacronym{stn}{STN}{Spatial Transformer Network}
\newacronym{svm}{SVM}{Support Vector Machine}
\newacronym{yolo}{YOLO}{You Only Look Once}

\newcommand{\dataset}{UFPR-AMR\xspace}
\begin{abstract}
In this paper, we tackle \gls*{amr} by leveraging the high capability of \glspl*{cnn}.
We design a two-stage approach that employs the Fast-YOLO object detector for counter detection and evaluates three different \gls*{cnn}-based approaches for counter recognition. 
In the \gls*{amr} literature, most datasets are not available to the research community since the images belong to a service company. 
In this sense, we introduce a new public dataset, called \dataset dataset, with 2,000 fully and manually annotated images. This dataset is, to the best of our knowledge, three times larger than the largest public dataset found in the literature and contains a well-defined evaluation protocol to assist the development and evaluation of \gls*{amr} methods.
Furthermore, we propose the use of a data augmentation technique to generate a balanced training set with many more examples to train the \gls*{cnn} models for counter recognition. In the proposed dataset, impressive results were obtained and a detailed speed/accuracy trade-off evaluation of each model was performed. In a public dataset, state-of-the-art results were achieved using less than 200 images for training.
\end{abstract}


{\noindent \footnotesize\textbf{*}Rayson Laroca,  \linkable{rblsantos@inf.ufpr.br}}

\begin{spacing}{2}   

\section{Introduction}
\label{sec:introduction}

\glsresetall

\gls*{amr} refers to automatically record the consumption of electric energy, gas and water for both monitoring and billing~\cite{shu2007study,gallo2015robust,gao2018automatic}. Despite the existence of smart readers~\cite{kabalci2016survey}, they are not widespread in many countries, especially in the underdeveloped ones, and the reading is still performed manually on site by an operator who takes a picture as reading proof~\cite{vanetti2013gas,gallo2015robust}. Since this operation is prone to errors, another operator needs to check the proof image to confirm the reading~\cite{vanetti2013gas,gallo2015robust}. This offline checking  is expensive in terms of human effort and time, and has low efficiency~\cite{cerman2016mobile}. 
Moreover, due to a large number of images to be evaluated, the inspection is usually done by sampling~\cite{quintanilha2017automatic} and errors might go unnoticed. 

Performing the meter inspection automatically would reduce mistakes introduced by the human factor and save manpower. 
Furthermore, the reading could also be executed fully automatically using cameras installed in the meter box~\cite{shu2007study,edward2013support}. Image-based \gls*{amr} has advantages such as lower cost and fast installation since it does not require renewal or replacement of existent meters~\cite{zhang2016automatic}.

A common \gls*{amr} approach includes three phases, namely: (i)~counter detection, (ii) digit segmentation and (iii) digit recognition. Counter detection is the fundamental stage, as its performance largely determines the overall accuracy and processing speed of the entire \gls*{amr} system. 

Despite the importance of a robust \gls*{amr} system
and that major advances have been achieved in computer vision using deep learning~\cite{lecun2015deep}, to the best of our knowledge, only in Ref.~\citenum{gomez2018cutting}, published very recently, \glspl*{cnn} were employed at all \gls*{amr} stages. Previous works relied, in at least one stage, on handcrafted features that capture certain morphological and color attributes of the meters/counters. These features are easily affected by noise and might not be robust to different types of meters.

Deep learning approaches are particularly dependent on the availability of large quantities of training data to generalize well and yield high classification accuracy for unseen data~\cite{salamon2017deep}. 
Some previous works~\cite{gallo2015robust,cerman2016mobile,gomez2018cutting} employed large datasets (e.g., more than $45$,$000$ images) to train and evaluate their systems. However, these datasets were not made public.
In the \gls*{amr} literature, the datasets are usually not publicly available since the images belong to the [electricity, gas, water] company.
In this sense, we introduce a new public dataset, called \dataset dataset, with $2$,$000$ fully annotated images to assist the development and evaluation of \gls*{amr} methods. The proposed dataset is three times larger than the largest public dataset~\cite{goncalves2016reconhecimento} found in the literature. 

In this paper, we design a two-stage approach for \gls*{amr}.
We first detect the counter region and then tackle the digit segmentation and recognition stages jointly by leveraging the high capability of \glspl*{cnn}. 
We employ a smaller version of the YOLO object detector, called Fast-YOLO~\cite{redmon2016yolo}, for counter detection. 
Afterward, we evaluate three CNN-based approaches, i.e. CR-NET~\cite{montazzolli2017}, Multi-Task Learning~\cite{goncalves2018realtime} and \gls*{crnn}~\cite{shi2017endtoend}, for the counter recognition stage (i.e., digit segmentation and recognition). CR-NET is a YOLO-based model proposed for license plate character detection and recognition, while Multi-Task and \gls*{crnn} are segmentation-free approaches designed respectively for the recognition of license plates and scene text. 
These approaches were chosen since promising results have been achieved through them in these applications. 
Finally, we propose the use of a data augmentation process to train the \gls*{cnn} models for counter recognition to explore different types of counter/digit deformations and their influence on the models' performance. 

The experimental evaluation demonstrates the effectiveness of the \gls*{cnn} models for \gls*{amr}. 
First, all counter regions were correctly located through Fast-YOLO in the proposed dataset and also in two public datasets found for this task~\cite{vanetti2013gas,goncalves2016reconhecimento}. Second, the CR-NET model yielded promising recognition results, outperforming both Multi-Task and \gls*{crnn} models in the \dataset dataset. 
Finally, an impressive recognition rate of \REV{$97.30$\%} was achieved using Fast-YOLO and CR-NET in a set of images proposed for end-to-end evaluations of \gls*{amr} systems, called Meter-Integration subset~\cite{vanetti2013gas}, against $85$\% and $87$\% achieved by the baselines~\cite{vanetti2013gas,gallo2015robust}. 
In addition, the CR-NET and Multi-Task models are able to achieve outstanding \gls*{fps} rates in a high-end GPU, being possible to process respectively $185$ and $250$ \gls*{fps}. 

Considering the aforementioned discussion, the main contributions of our work are summarized as follows:
\begin{itemize}
\item A two-stage \gls*{amr} approach with \glspl*{cnn} being employed for both counter detection and recognition. In the latter, three different types of \gls*{cnn} are evaluated; 
\item A public dataset for \gls*{amr} with $2$,$000$ fully and manually annotated images/meters (i.e.,~$10$,$000$ digits) with a well-defined evaluation protocol, allowing a fair comparison between different approaches for this task;
\item The \gls*{cnn}-based approaches outperformed all baselines in public datasets and achieved impressive results in both accuracy and computational time in the proposed \dataset dataset.
\end{itemize}

The remainder of this paper is organized as follows. We briefly review related works in Section~\ref{sec:related_work}. The \dataset dataset is introduced in Section~\ref{sec:dataset}. The methodology is presented in Section~\ref{sec:proposed}. We report and discuss the results in Section~\ref{sec:results}. Conclusions and future work are given in Section~\ref{sec:conclusion}.
\section{Related Work}
\label{sec:related_work}

\gls*{amr} intersects with other \gls*{ocr} applications, such as license plate recognition~\cite{du2013review} and robust reading~\cite{karatzas2015icdar}, as it must reliably extract text information from images taken under different conditions. Although \gls*{amr} is not as widespread in the literature as these applications, a satisfactory number of works have been produced in recent years~\cite{anis2017digital,gomez2018cutting,cerman2016mobile,gao2018automatic,quintanilha2017automatic}.  
Here, we briefly survey these works by first describing the approaches employed for each \gls*{amr} stage. Next, we present some papers that address two stages jointly or using the same method. Then, we discuss the deep learning approaches and datasets used so far. Finally, we conclude this section with final remarks.

\vspace{1mm}
\noindent \textbf{Counter Detection}:
Many pioneering approaches exploited the vertical and horizontal pixel projections histograms for counter detection~\cite{zhao2005research,shu2007study,edward2013support}. Projection-based methods can be easily affected by the rotation of the counter. Refs.~\citenum{elrefaei2015automatic,quintanilha2017automatic,anis2017digital,gallo2015robust,cerman2016mobile,goncalves2016reconhecimento} took advantage of prior knowledge such as counter's position and/or its colors (e.g., green background and red decimal digits). 
A major drawback of these techniques is that they might not work on all meter types and the color information might  not be stable when the illumination changes. Other works include the use of template matching~\cite{quintanilha2017automatic} and the AdaBoost classifier~\cite{gao2018automatic}. In the latter, normalized gradient magnitude, \gls*{hog} and LUV color channels were adopted as low-level feature descriptors.

\vspace{1mm}
\noindent \textbf{Digit Segmentation}:
Projection and color-based approaches have also been widely employed for digit segmentation~\cite{elrefaei2015automatic,zhang2016automatic,rodriguez2014hdmr}. The use of morphological operations with \gls*{cca} was considered in Refs.~\citenum{cerman2016mobile,anis2017digital}. However, it presents the drawback of depending  largely on the result of binarization as it cannot segment digits correctly if they are connected or broken. In Ref.~\citenum{edward2013support}, a binary digit/non-digit \gls*{svm} was applied in a sliding window fashion, while Gallo et al.~\cite{gallo2015robust} exploited \gls*{mser}. In Ref.~\citenum{gallo2015robust}, the \gls*{mser} algorithm failed to segment digits in images with problems such as bluring and perspective distortions.

\vspace{1mm}
\noindent \textbf{Digit Recognition}:
Template matching~\cite{elrefaei2015automatic,rodriguez2014hdmr,zhao2005research} along with simple measures of similarity have been widely used for digit recognition. Nevertheless, it is known that if a digit is different from the template due to any font change, rotation or noise, this approach produces incorrect recognition~\cite{du2013review}. 
Thus, many authors have employed an \gls*{svm} classifier for digit recognition.
In Refs.~\citenum{edward2013support,vanetti2013gas}, simple features such as pixel intensity were used in training, while \gls*{hog} descriptors were adopted as features in Refs.~\citenum{quintanilha2017automatic,gallo2015robust}. Although some promising results have been attained, it should be noted that it is not trivial to find the appropriate hyper-parameters of \gls*{svm} classifiers as well as the best features to be extracted. The open-source Tesseract \gls*{ocr} Engine~\cite{smith2007overview} was applied in Refs.~\citenum{nodari2011multineural,cerman2016mobile,vanetti2013gas}, however, satisfactory results were not obtained in any of them. 
Cerman et al.~\cite{cerman2016mobile} achieved a remarkable improvement in digit recognition when using a \gls*{cnn} inspired by the LeNet-$5$ architecture instead of Tesseract.  

\gls*{amr} presents an unusual challenge in \gls*{ocr}: rotating digits. Typically, this is the major cause of errors, even when robust approaches are employed for digit recognition~\cite{zhao2009design,gao2018automatic}. In Ref.~\citenum{rodriguez2014hdmr}, this problem was addressed using a Hausdorff distance based algorithm, achieving excellent recognition results in real time. Note that all images were extracted from a single meter and, as pointed out by the authors, a controlled environment was required since there were no preprocessing stage and no algorithm for angle correction.

\vspace{1mm}
\noindent \textbf{Miscellaneous}: Nodari \& Gallo~\cite{nodari2011multineural} exploited an ensemble of \gls*{mlp} networks to perform the counter detection and digit segmentation without preprocessing and postprocessing stages. Since low F-measure rates were achieved, extra techniques were added in Ref.~\citenum{vanetti2013gas}, an extension of Ref.~\citenum{nodari2011multineural}.
In summary, a watershed algorithm was applied to improve counter detection and Fourier analysis was employed to avoid false positives in digit segmentation. 
Although better results were attained, only $100$ images were used to evaluate their system performance, which may not be representative enough. It should be noted that, to the best of our knowledge, this was the first work to make the images used in the experiments publicly available. 

Gao et al.~\cite{gao2018automatic} designed a bidirectional \gls*{lstm} network for counter recognition. 
In their approach, a feature sequence is first generated by a network that combines convolutional and recurrent layers. Then, an attention decoder predicts, recurrently, one digit at each step according to the feature representation. 
A promising accuracy rate was reported, with most of the errors appearing in cases of half digits.

Gómez et al.~\cite{gomez2018cutting} presented a segmentation-free \gls*{amr} system able to output readings directly without explicit counter detection. A \gls*{cnn} architecture was trained in an end-to-end manner where the initial convolutional layers extract visual features of the whole image and the fully connected layers predict the probabilities for each digit. 
Even though an impressive overall accuracy was achieved, their approach was evaluated only on a large private dataset which has almost $180k$ training samples and mostly images with the counter well centered and occupying a good portion of the image. 
Thus, as pointed out by the authors, small-meter images pose difficulties to their system.

\vspace{1mm}
\noindent \textbf{Datasets}:
To the best of our knowledge, only Refs.~\citenum{goncalves2016reconhecimento,vanetti2013gas} made available the datasets used in their experiments. These datasets are composed of gas meter images with resolution of $640\times480$ pixels (mostly) and the counter occupying a large portion of the image, which facilitates its detection. Additionally, both datasets are small ($253$ and $640$ images, respectively) and the cameras used to capture them were not specified. It is important to note that in the dataset introduced in Ref.~\citenum{vanetti2013gas}, $153$ images are divided into different subsets for the evaluation of each stage and only $100$ images are for the end-to-end evaluation of the \gls*{amr} system. Also, there is no split protocol in Ref.~\citenum{goncalves2016reconhecimento}, which prevents a fair comparison between different approaches.

\vspace{1mm}
\noindent \textbf{Deep Learning}:
Recently, deep learning approaches have won many machine learning competitions and challenges, even achieving superhuman visual results in some domains~\cite{schmidhuber2015deep}. Such a fact motivated us to employ deep learning for \gls*{amr}, since we could find only three works~\cite{cerman2016mobile,gao2018automatic,gomez2018cutting} employing \glspl*{cnn} in this context and all of them made use of large private datasets, overlooking the public datasets. This suggests that these models are able to generalize only with many training samples (e.g., $177$,$758$ images in the segmentation-free system proposed in Ref.~\citenum{gomez2018cutting}). Moreover, (i) conventional image processing with handcrafted features was used in at least one stage in Refs.~\citenum{cerman2016mobile,gao2018automatic}, (ii)~the images used in Ref.~\citenum{gao2018automatic} are mostly sharp and very similar, which does not represent real-world conditions, and~(iii) the poor digit segmentation accuracy obtained in Ref.~\citenum{cerman2016mobile}, i.e. $81$\%, through a sequence of conventional image processing methods, discourages its use in real-world applications.

\vspace{1mm}
\noindent \textbf{Final Remarks:} The approaches developed for \gls*{amr} are still limited. In addition to the aforementioned points (i.e., private datasets and handcrafted features), many authors do not report the computational time of their approaches, making it difficult an accurate analysis of their speed/accuracy trade-off, as well as their applicability. In this paper, \glspl*{cnn} are used for both counter detection and recognition. 
We evaluate the \glspl*{cnn} that achieved state-of-the-art results in other applications in both the proposed and public datasets, reporting the accuracy and the computational time to enable fair comparisons in future works. 
\section{The \dataset Dataset}
\label{sec:dataset}

\begin{figure*}[!htb]
	\begin{center}

	\includegraphics[width=0.97\columnwidth]{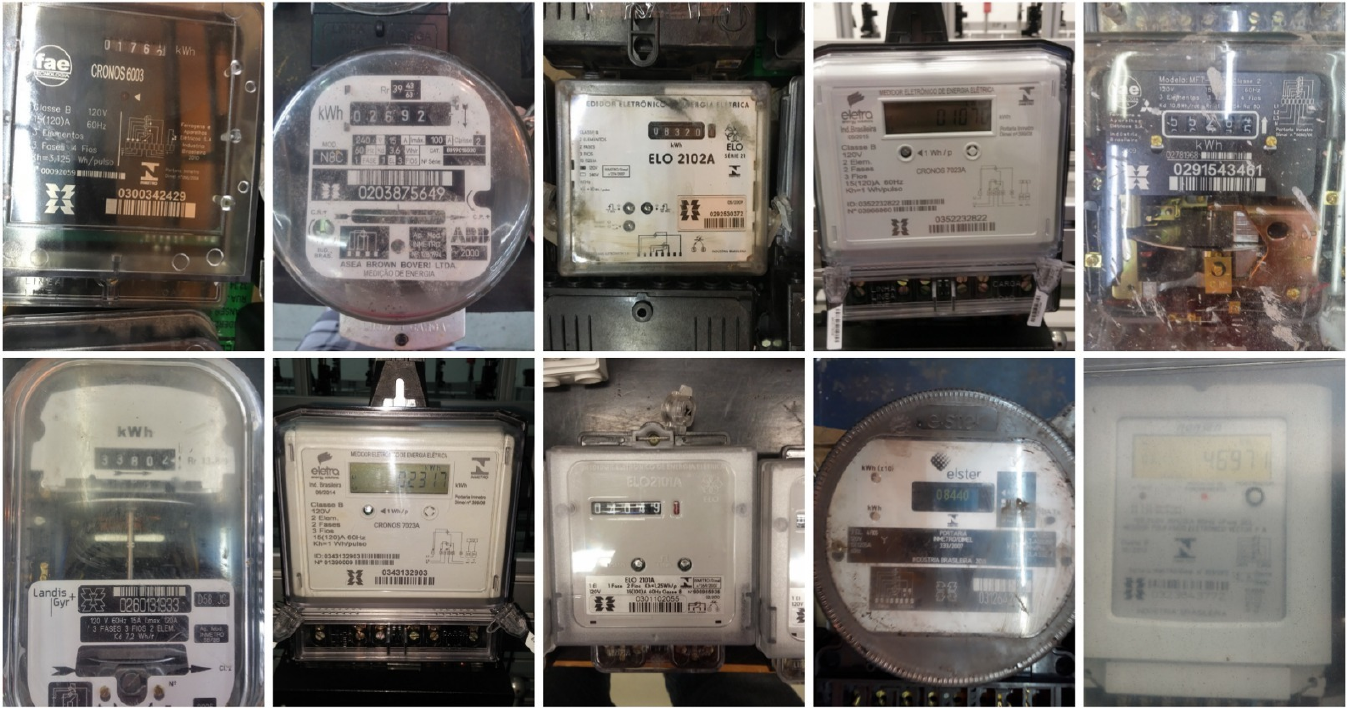}
	\end{center}
	\vspace{-2mm}
	\caption{Sample images of the \dataset dataset (some images were slightly resized for display purposes). Note the diversity of meter types and conditions, as well as the existence of several textual blocks similar to the counter region.}
	
	\label{fig:dataset}    
\end{figure*}

The proposed dataset contains $2$,$000$ images taken from inside a warehouse of the \gls*{copel}, which directly serves more than $4$ million consuming units in the Brazilian state of Paraná~\cite{copel}. Therefore, our dataset presents electric meters of different types and in different conditions. The diversity of the dataset is shown in Fig.~\ref{fig:dataset}. One can see that (i) the counter occupies a small portion in the image, which makes its location more difficult; (ii) there are several similar textual blocks (e.g., meter specifications and serial number) to the counter region. The \dataset dataset is publicly available to the research community at \small\url{https://web.inf.ufpr.br/vri/databases/ufpr-amr/}\normalsize.

Meter images commonly have some artifacts (e.g., blur, reflections, low contrast, broken glass, dirt, among others) due to the meter's conditions and the misuse of the camera by the human operator, which may impair the reading of electric energy consumption. 
In addition, it is possible that the digits are rotating or in-between positions (e.g., a digit going from $4$ to $5$) in some types of counters. In such cases, we consider the lowest digit as the ground truth, since this is the protocol adopted at \gls*{copel}. The exception, to have a reasonable rule, is between digits $9$ and $0$, where it should be labeled as~$9$. 

The images were acquired with three different cameras and are available in the JPG format with resolution between $2$,$340\times4$,$160$ and $3$,$120\times4$,$160$ pixels. 
The cameras used were: \textit{LG G3 D855}, \textit{Samsung Galaxy J7 Prime} and \textit{iPhone~6s}. As the cameras (cell phones) belong to different price ranges, the images presumably have different levels of quality. Additional information can be seen in Table~\ref{tab:dataset_info}.

\vspace{3mm}
\begin{table}[!htb]
    \caption{Additional information about the \dataset dataset: (a) how many images were captured with each camera; (b) dimensions of counters and digits (width $\times$ height in pixels). It is noteworthy the large variation in the sizes of counters and digits.}%
    \label{tab:dataset_info}%
    \vspace{-5mm}
	\begin{center}
	\resizebox{!}{21.3pt}{
		\subfloat[][]{\begin{tabular}{@{}lc@{}}
			\toprule
			\textbf{Camera}     & \textbf{Images} \\ \midrule
			\textit{LG G3} & $947$           \\
			\textit{J7 Prime}   & $584$           \\ 
			\textit{iPhone 6s}  & $469$           \\ \midrule
			Total               & $2$,$000$          \\ \bottomrule
			\end{tabular}}}%
	\qquad
	\resizebox{!}{23pt}{
		\subfloat[][]{\begin{tabular}{@{}lcc@{}}
			\toprule
			\textbf{Info} & \textbf{Counters} & \textbf{Digits} \\ \midrule
			Minimum Size & $238\times96$ & $35\times63$ \\
			Maximum Size & $1{,}689\times365$ & $168\times283$ \\
			Average Size & $674\times178$ & $76\times134$ \\
			Aspect Ratio & $3.79$ & $0.57$ \\ \bottomrule
			\end{tabular}}}
	\end{center}
	\vspace{-6mm}
\end{table}

Every image has the following annotations available in a text file: the camera in which the image was taken, the counter position $(x,y,w,h)$, the reading, as well as the position of each digit. 
All counters of the dataset (regardless of meter type) have $5$ digits, and thus $10$,$000$ digits were manually annotated.

\begin{figure}[!htb]
	\begin{center}
	\includegraphics[width=0.75\columnwidth]{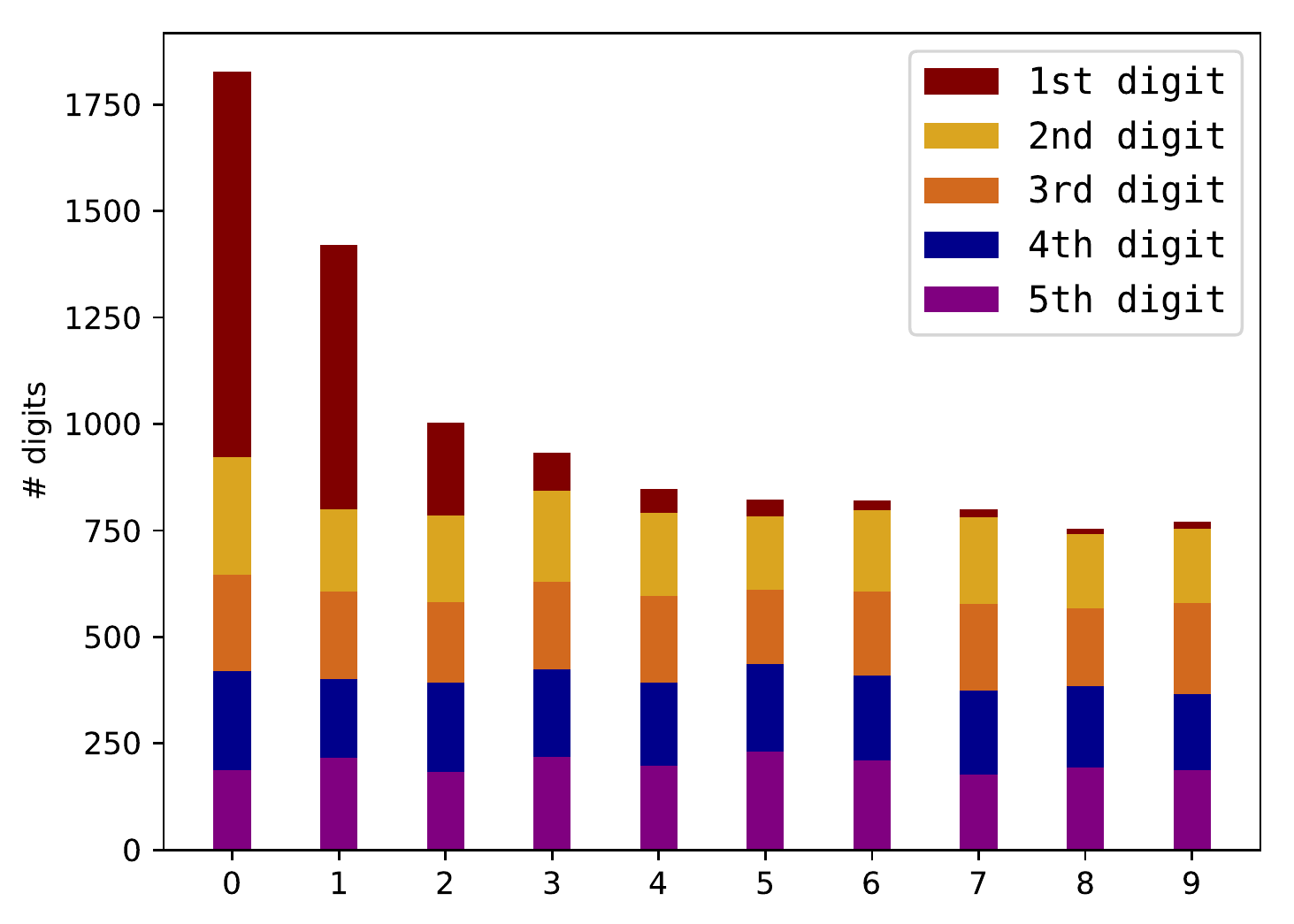}
	\end{center}
	\vspace{-3mm}
	\caption{Frequency distribution of digits in the \dataset dataset. It is worth noting that the first position (i.e., the most significant) consists almost exclusively of $0$s and $1$s. On the other hand, the frequency of digits in the other positions is very well balanced.}
	\label{fig:frequency}   
\end{figure}

Remark that a brand new meter starts with \texttt{00000} and the most significant digit positions take longer to be increased.
Then, it is natural that the less significant digits (i.e.,~$0$ and~$1$) have many more instances than the others. 
Nonetheless, digits $4$-$9$ have a fairly similar number of examples. Fig.~\ref{fig:frequency} shows the distribution of the digits in the \dataset dataset.  

The dataset is split into three sets: training ($800$~images), validation ($400$~images) and test ($800$~images).
We adopt this protocol (i.e., with a larger test set) since it has already been adopted in other datasets~\cite{goncalves2016benchmark,laroca2018robust} and to provide more samples for analysis of statistical significance. \REV{It should be noted that this division was made randomly and the sets generated are explicitly available along with the \dataset dataset. Additionally, experiments carried out by us suggested that dividing the dataset multiple times and then averaging the results is not necessary, as the proposed division is representative.}
\section{Methodology}
\label{sec:proposed}

Meters have many textual blocks that can be confused with the counter's reading. Moreover, the \gls*{roi} (i.e., the counter) usually occupies a small portion of the image and its position varies according to the meter type. Therefore, we propose to first locate the counter region and then perform its recognition in the detected patch. 
We tackle both stages by leveraging the high capability of state-of-the-art \glspl*{cnn}.
It is remarkable that, to the best of our knowledge, this is only the second work in which both stages are addressed using \glspl*{cnn}~\cite{gomez2018cutting} and the first with the experiments being performed on public datasets. 

In the following sections, we describe the \gls*{cnn} models employed for counter detection and counter recognition.
It is worth noting that all parameters (e.g., \glspl*{cnn} input size, number of epochs, among others) specified here are defined based on the validation set and presented in Section~\ref{sec:results}, where the experiments are reported. 

\subsection{Counter Detection}
\label{subsec:proposed_detection}

Recently, great progress has been made in object detection through models inspired by YOLO~\cite{redmon2016yolo,wu2017squeezedet,tripathi2017lcdet}, a \gls*{cnn}-based object detection system that (i) reframes object detection as a single regression problem; (ii) achieved outstanding and state-of-the-art results in the PASCAL VOC and COCO detection tasks~\cite{redmon2017yolo9000}.
For that reason, we decided to fine-tune it for counter detection. However, as we want to detect only one class and the computational cost is one of our main concerns, we chose to use a smaller model, called Fast-YOLO~\cite{redmon2016yolo}, which uses fewer convolutional layers than YOLO and fewer filters in those layers. Despite being smaller, Fast-YOLO (architecture shown in Table~\ref{tab:fast_yolo2}) yielded outstanding  results, i.e. detections with \gls*{iou}~$\ge 0.8$ with the ground truth, in preliminary experiments.
\REV{The \gls*{iou} is often used to assess the quality of predictions in object detection tasks~\cite{everingham2010pascalvoc} and can be expressed by the formula
\vspace{-3mm}
\begin{equation}
    \textit{\gls*{iou}} = \frac{\textit{area}(B_p \cap B_{gt})}{\textit{area}(B_p \cup B_{gt})} \, ,
\end{equation}
\vspace{-13mm}

\noindent where $B_{p}$ and $B_{gt}$ are the predicted and ground truth bounding boxes, respectively. The closer the \gls*{iou} is to $1$, the better the detection.}
For this reason, we believe that very deep models are not necessary to handle the detection of a single class of objects.

\begin{table}[!htb]
    \caption{Fast-YOLO network used to detect the counter region.}
    \vspace{-2mm}
	\begin{center}
	\label{tab:fast_yolo2}	
	\resizebox{0.65\columnwidth}{!}{	
		\begin{tabular}{@{}cccccc@{}}
			\toprule
\multicolumn{2}{c}{\textbf{Layer}} & \textbf{Filters} & \textbf{Size} & \textbf{Input} & \textbf{Output} \\ \midrule
$0$ & conv & $16$ & $3 \times 3 / 1$ & $416 \times 416 \times 3$ & $416 \times 416 \times 16$ \\
$1$ & max &  & $2 \times 2 / 2$ & $416 \times 416 \times 16$ & $208 \times 208 \times 16$ \\
$2$ & conv & $32$ & $3 \times 3 / 1$ & $208 \times 208 \times 16$ & $208 \times 208 \times 32$ \\
$3$ & max &  & $2 \times 2 / 2$ & $208 \times 208 \times 32$ & $104 \times 104 \times 32$ \\
$4$ & conv & $64$ & $3 \times 3 / 1$ & $104 \times 104 \times 32$ & $104 \times 104 \times 64$ \\
$5$ & max &  & $2 \times 2 / 2$ & $104 \times 104 \times 64$ & $52 \times 52 \times 64$ \\
$6$ & conv & $128$ & $3 \times 3 / 1$ & $52\times 52 \times 64$ & $52 \times 52 \times 128$ \\
$7$ & max &  & $2 \times 2 / 2$ & $52 \times 52 \times 128$ & $26 \times 26 \times 128$ \\
$8$ & conv & $256$ & $3 \times 3 / 1$ & $26 \times 26 \times 128$ & $26 \times 26 \times 256$ \\
$9$ & max &  & $2 \times 2 / 2$ & $26 \times 26 \times 256$ & $13 \times 13 \times 256$ \\
$10$ & conv & $512$ & $3 \times 3 / 1$ & $13 \times 13 \times 256$ & $13 \times 13 \times 512$ \\
$11$ & max &  & $2 \times 2 / 1$ & $13 \times 13 \times 512$ & $13 \times 13 \times 512$ \\
$12$ & conv & $1024$ & $3 \times 3 / 1$ & $13 \times 13 \times 512$ & $13 \times 13 \times 1024$ \\
$13$ & conv & $1024$ & $3 \times 3 / 1$ & $13 \times 13 \times 1024$ & $13 \times 13 \times 1024$ \\
$14$ & conv & $30$ & $1 \times 1 / 1$ & $13 \times 13 \times 1024$ & $13 \times 13 \times 30$ \\
$15$ & detection &  &  &  &  \\ \bottomrule
		\end{tabular}}
	\end{center}
\end{table}

For counter detection, we use the weights pre-trained on ImageNet~\cite{imagenet2009} and perform two minor changes in the Fast-YOLO model. 
First, we recalculate the anchor boxes for the \dataset dataset using the algorithm available in Ref.~\citenum{alexeyab}. \REV{Anchors are initial shapes that serve as references at multiple scales and aspect ratios. Instead of predicting arbitrary bounding boxes, YOLO only adjusts the size of the nearest anchor to the size of the object. Predicting offsets instead of coordinates simplifies the problem and makes it easier for the network to learn~\cite{redmon2017yolo9000}.}
Then, we reduce the number of filters in the last convolutional layer from $125$ to $30$ to output $1$ class instead of $20$. The number of filters in the last layer is given by
\vspace{-1mm}
\begin{equation} 
\label{eq:filters}
\textit{filters} = (C + 5) \times A \, ,
\end{equation} 
\noindent where $A$ is the number of anchor boxes (we use $A$~=~$5$) used to predict bounding boxes. 
Each bounding box has four coordinates $(x, y, w, h)$, a objectness value~\cite{alexe2012measuring} \REV{(i.e., how likely the bounding box contains an object) along with the probability of that object belonging to each of the $C$ classes, in our case $C=1$ (i.e., only the counter region)~\cite{redmon2017yolo9000}.}
Remark that the choice of appropriate anchor boxes is very important, and thus our boxes are similar to counters in size and aspect ratio.

We employ Fast-YOLO's multi-scale training~\cite{redmon2017yolo9000}. In short, every $10$ batches, the network randomly chooses a new image dimension size from $320\times320$ to $608\times608$ pixels (default values). These dimensions were chosen considering that the Fast-YOLO model down samples the image by a factor of $32$. As pointed out in Ref.~\citenum{redmon2017yolo9000}, this approach forces the network to learn to predict well across a variety of input dimensions.
Then, we use $416\times416$ images as input since the best results (speed/accuracy trade-off in the validation set) were obtained with this dimension as input. It is remarkable that, although YOLO networks have a $1:1$ input aspect ratio, previous works~\cite{laroca2018robust,montazzolli2017} have attained excellent object detection results (over $99$\% recall) in images with different aspect ratios (e.g.,~$1$,$920\times1$,$080$). \REV{All image resizing operations were performed using bilinear interpolation.}

In cases where more than one counter is detected, we consider only the detection with the highest confidence since each image/meter has only one counter. To avoid losing digits in cases where the counter is not very well detected, we add a margin (with size chosen based on the validation set) on the detected patch so that all digits are within it for the recognition stage. A negative recognition result is given in cases where no counter is found.

\subsection{Counter Recognition}
\label{subsec:proposed_recognition}

We employ three \gls*{cnn}-based approaches for performing counter recognition: CR-NET~\cite{montazzolli2017}, Multi-Task Learning~\cite{goncalves2018realtime} and \gls*{crnn}~\cite{shi2017endtoend}. These models were chosen because promising results were obtained through them in other \gls*{ocr} applications, such as license plate recognition and scene text recognition. It is noteworthy that, unlike CR-NET, the last two models do not need the coordinates of each digit in the training phase.
In other words, Multi-Task Learning and \gls*{crnn} approaches only need the counter's reading.
This is of paramount importance in cases where a large number of images  is available for learning (e.g., millions or hundreds of thousands), since manually labeling each digit is very costly and prone to errors.

The remainder of this section is organized into four parts, one to describe the data augmentation method, which is essential to effectively train the deep models, and one part for each \gls*{cnn} approach employed for counter recognition. 

\subsubsection{Data Augmentation}
\label{subsubsec:data_augmentation}

It is well known that unbalanced data is undesirable for neural network classifiers since the learning of some patterns might be biased. For instance, some classifiers may learn to always classify the first digit as $0$, but this is not always the case (see Fig.~\ref{fig:frequency}), although it is by far the most common. 
To address this issue, we employ the data augmentation technique proposed in Ref.~\citenum{goncalves2018realtime}. Using this technique, we are able to create a new set of images, where each digit class is equally represented in every position.
This set consists of permutations of the original images.
The order and frequency of the digits in the generated counters are chosen to uniformly distribute the digits along the positions. Note that the location of each digit (i.e., its bounding box) is required to apply this data augmentation technique.

\begin{figure}[!htb]
	\centering
	
	\includegraphics[width=0.78\columnwidth]{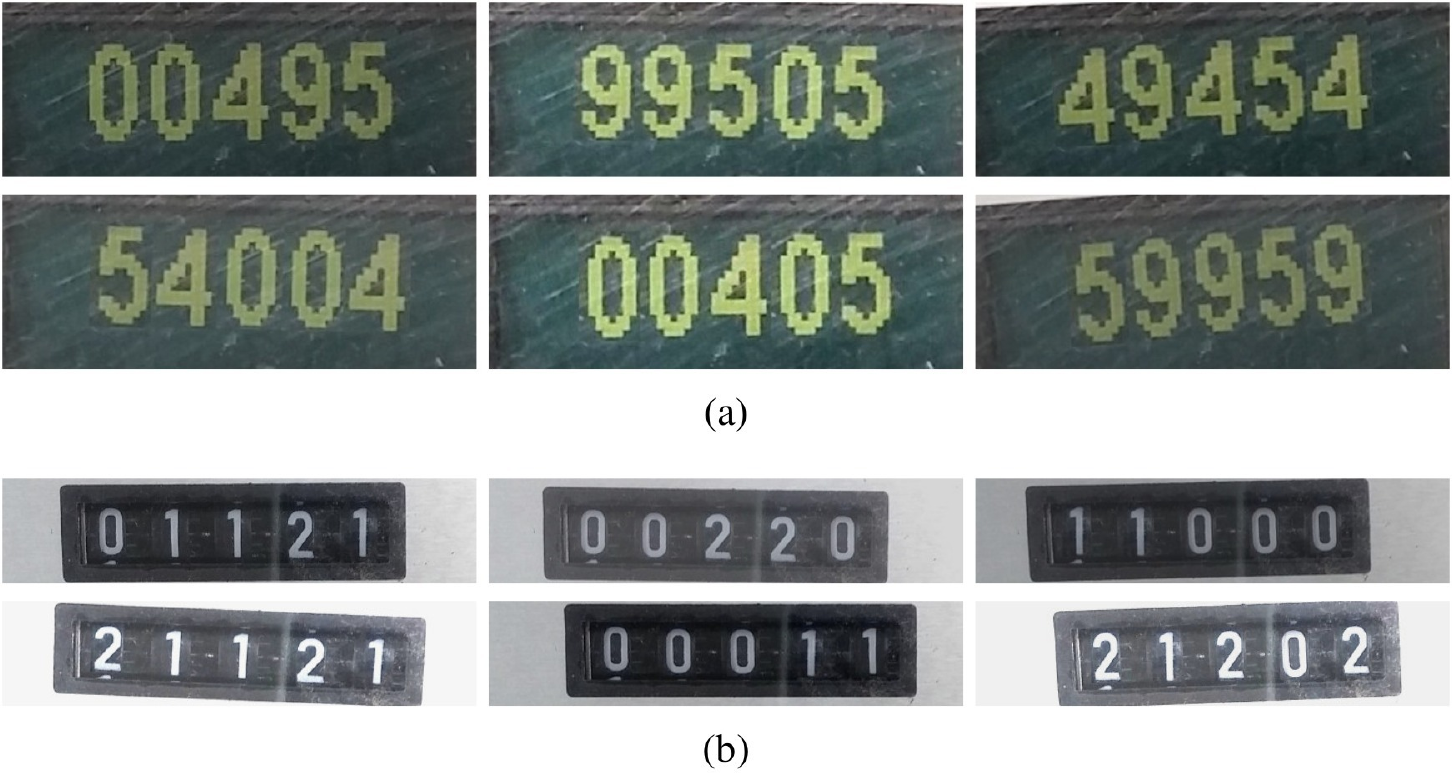}
	\vspace{2mm}
	
	\caption{Data augmentation examples, where the images in the upper-left corner of (a) and (b) are the originals, and the others were generated automatically. In (a) and (b), counters of different types and aspect ratios are shown.}
	\label{fig:data_augmentation}    
\end{figure}

Some artificially generated images when applying the method in the \dataset dataset are shown in Fig.~\ref{fig:data_augmentation}.
We also perform random variations of brightness, rotation and crop coordinates to increase even more the robustness of our augmented images, creating new training examples for the~\glspl*{cnn}. As can be seen, the data augmentation approach works on different types of meters.

\captionsetup[subfigure]{position=bottom}

The adjustment of parameters is of paramount importance for the effectiveness of this technique since the presence of very large variations in brightness, rotation or cropping, for instance, might impair the recognition through the generation of images that do not match real scenarios. 
\REV{Therefore, the parameter ranges were empirically determined based on experiments performed on the validation set, i.e., brightness variation of the pixels $[0.5; 2]$, rotation angles between \minus$5\degree$~and~$5\degree$ and cropping from \minus$2$\%~to~$8$\% of the counter size.
Once these ranges were established, new counter images were generated using random values within those ranges for each parameter.}

\subsubsection{CR-NET}
\label{subsubsec:crnet}

CR-NET is a YOLO-based model proposed for license plate character detection and recognition~\cite{montazzolli2017}. This model consists of the first eleven layers of YOLO and four other convolutional layers added to improve non-linearity. In Ref.~\citenum{montazzolli2017}, CR-NET (with an input size of $240\times80$ pixels) was capable of detecting and recognizing license plate characters at $448$ \gls*{fps}. Laroca et al.~\cite{laroca2018robust} also achieved great results applying CR-NET for this purpose.

The CR-NET architecture is shown in Table~\ref{tab:yolo_montazzolli}. As in the counter detection stage, we recalculate the anchors for our data and make adjustments in the number of filters in the last layer.
Furthermore, we adapt the input image size taking into account the aspect ratio of the counters, which have a different aspect ratio when compared to license plates in Ref.~\citenum{montazzolli2017}.
Then, we use as input an image with resolution of $400\times106$ pixels since the results obtained when using other sizes (e.g., $360\times95$ and $440\times116$) were worse or similar, but with a higher computational~cost. 

\begin{table}[!htb]
\caption{CR-NET with some modifications for counter recognition: input size of $400\times106$ pixels and $75$ filters in the last layer.}
\vspace{-2mm}
\label{tab:yolo_montazzolli}
\begin{center}
\resizebox{0.65\columnwidth}{!}{
\begin{tabular}{@{}cccccc@{}}
\toprule
\multicolumn{2}{c}{\textbf{Layer}} & \textbf{Filters} & \textbf{Size} & \textbf{Input} & \textbf{Output} \\ \midrule
$0$ & conv & $32$ & $3 \times 3 / 1$ & $400 \times 106 \times 3$ & $400 \times 106 \times 32$ \\
$1$ & max &  & $2 \times 2 / 2$ & $400 \times 106 \times 32$ & $200 \times 53 \times 32$ \\
$2$ & conv & $64$ & $3 \times 3 / 1$ & $200 \times 53 \times 32$ & $200 \times 53 \times 64$ \\
$3$ & max &  & $2 \times 2 / 2$ & $200 \times 53 \times 64$ & $100 \times 26 \times 64$ \\
$4$ & conv & $128$ & $3 \times 3 / 1$ & $100 \times 26 \times 64$ & $100 \times 26 \times 128$ \\
$5$ & conv & $64$ & $1 \times 1 / 1$ & $100 \times 26 \times 128$ & $100 \times 26 \times 64$ \\
$6$ & conv & $128$ & $3 \times 3 / 1$ & $100\times 26 \times 64$ & $100 \times 26 \times 128$ \\
$7$ & max & & $2 \times 2 / 2$ & $100 \times 26 \times 128$ & $50 \times 13 \times 128$ \\
$8$ & conv & $256$ & $3 \times 3 / 1$ & $50 \times 13 \times 128$ & $50 \times 13 \times 256$ \\
$9$ & conv & $128$ & $1 \times 1 / 1$ & $50 \times 13 \times 256$ & $50 \times 13 \times 128$ \\
$10$ & conv & $256$ & $3 \times 3 / 1$ & $50 \times 13 \times 128$ & $50 \times 13 \times 256$ \\
$11$ & conv & $512$ & $3 \times 3 / 1$ & $50 \times 13 \times 256$ & $50 \times 13 \times 512$ \\
$12$ & conv & $256$ & $1 \times 1 / 1$ & $50 \times 13 \times 512$ & $50 \times 13 \times 256$ \\
$13$ & conv & $512$ & $3 \times 3 / 1$ & $50 \times 13 \times 256$ & $50 \times 13 \times 512$ \\
$14$ & conv & $75$ & $1 \times 1 / 1$ & $50 \times 13 \times 512$ & $50 \times 13 \times 75$ \\
$15$ & detection &  &  &  &  \\ \bottomrule
\end{tabular}}
\end{center}
\end{table}

\begin{figure}[!htb]
	\begin{center}
	{\includegraphics[width=0.5\columnwidth]{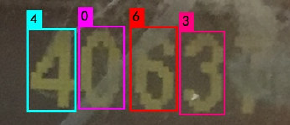}}
    \end{center}
    \vspace{-1mm}
    \caption{A counter where less than $5$ digits were detected/recognized by the CR-NET. We could employ leading zeros (e.g.,~$4063 \rightarrow 04063$), however, this could result in a large error in the meter reading.}
	\label{fig:leading_zeros}   
\end{figure}

We consider only the five digits detected/recognized with highest confidence, since commonly more than five digits are predicted. However, we noticed that the same digit might be detected more than once by the network. Therefore, we first apply a non-maximal suppression algorithm to eliminate redundant detections. Although \emph{highly unlikely} (i.e., $\approx0.1$\%), it is also possible that less than five digits are detected by the CR-NET, as shown in Fig~\ref{fig:leading_zeros}. In such cases, we reject the counter's recognition. 

\subsubsection{Multi-Task Learning}
\label{subsubsec:multitask}

Multi-Task Learning is another approach for character string recognition developed for license plates~\cite{goncalves2018realtime, jakub2017holistic}.
This method skips the character segmentation stage and directly recognizes the character string of an image (here, the cropped counter).
Since there might be multiple characters, each character is modeled as a task on the network.

For the \dataset dataset, we use a similar architecture adding the restraint that each character must be a digit, transforming the output space from $36$ (their work considers numbers and letters) to $10$ for each digit.
The architecture holistically segments and recognizes all five characters due to its multi-task output.

Table~\ref{tab:multitask_model} shows the architecture of the model, which is very compact with only $4$ convolutional layers followed by a fully connected shared layer and two fully connected layers for each digit, indexed from $1$ to $5$. 
Each output represents the classification of one of the digits. Thus, no explicit segmentation is performed in this approach.

\begin{table}[!htb]
\caption{Multi-Task layers and hyperparameters.}
\label{tab:multitask_model}
\vspace{-2mm}
\begin{center}
\resizebox{0.65\columnwidth}{!}{
\begin{tabular}{@{}cccccc@{}}
\toprule
\multicolumn{2}{c}{\textbf{Layer}} & \textbf{Filters} & \textbf{Size} & \textbf{Input} & \textbf{Output} \\ \midrule
        $0$ & conv & $128$ & \REV{$5 \times 5 / 1$} & $220 \times 60 \times 1$ & $220 \times 60 \times 128$ \\
        $1$ & max & & $2 \times 2 / 2$ & $220 \times 60 \times 128$ & $110 \times 30 \times 128$ \\
        $2$ & conv & $128$ & $3 \times 3 / 1$ & $110 \times 30 \times 128$ & $110 \times 30 \times 128$ \\
        $3$ & conv & $192$ & $3 \times 3 / 1$ & $110 \times 30 \times 128$ & $110 \times 30 \times 192$ \\
        $4$ & max & & $2 \times 2 / 2$ & $110 \times 30 \times 192$ & $55 \times 15 \times 192$ \\
        $5$ & conv & $256$ & $3 \times 3 / 1$ & $55 \times 15 \times 192$ & $55 \times 15 \times 256$ \\
        $6$ & max & & $2 \times 2 / 2$ & $55 \times 15 \times 256$ & $27 \times 7 \times 256$ \\
        $7$ & flatten & $ $ &  & $27 \times 7 \times 256$ & $48384$ \\[3pt] 
        \toprule
        
        \multicolumn{2}{c}{\textbf{Layer}} & \multicolumn{2}{c}{\REV{\textbf{Neurons}}} & \textbf{Input} & \textbf{Output} \\ \midrule
        
        $8$ & dense & \multicolumn{2}{c}{$4096$} & $48384$ & $4096$ \\
        $9$ & dense[$1..5$] & \multicolumn{2}{c}{$512$} & $4096$ & $512$ \\
        $10$ & dense[$1..5$] & \multicolumn{2}{c}{$10$} & $512$  & $10$ \\
        \bottomrule
    \end{tabular}}
\end{center}
\end{table}

\subsubsection{Convolutional Recurrent Neural Network}
\label{subsubsec:crnn}

\gls*{crnn}~\cite{shi2017endtoend} is a model designed for scene text recognition that consists of convolutional layers followed by recurrent layers, in addition to a custom transcription layer to convert the per-frame predictions into a label sequence. 
Given the counter patch, containing the digits, the convolutional layers act as a feature extractor, which is then transformed into a sequence of feature vectors and fed into an \gls*{lstm}~\cite{gers1997learning} recurrent layer. 
This layer handles the input as a sequence labeling problem, predicting a label distribution $y = {y_1, y_2,..., y_t}$ for each feature vector $x = {x_1, x_2,...,x_t}$ from the feature map.

The \gls*{ctc}~\cite{graves2006connectionist} cost function is adopted for sequence decoding. The \gls*{ctc} has a softmax layer with a label more than the original $10$ digits. The activation of each feature vector corresponds to a unique label that can be one of the ten digits or a `blank' (i.e., the absence of digit). 
\REV{Thus, this model is able to predict a variable number of digits, differently from Multi-Task where $5$ digits are always predicted.} As the classification is done through the whole feature map from the convolutional layers, digit segmentation is not required. 

We evaluate different network architectures with variations in the input size and in the number of filters and convolutional layers. As shown in Table~\ref{tab:crnn_model}, the input size is $160\times40$ pixels and there are only one \gls*{lstm} layer (instead of two, as in Ref.~\citenum{shi2017endtoend}) since the best results (considering the speed/accuracy trade-off) in the validation set were obtained with these parameters.

\begin{table}[!htb]
\caption{\gls*{crnn} layers and hyperparameters.}
\vspace{-2mm}
\label{tab:crnn_model}
\begin{center}
\resizebox{0.60\columnwidth}{!}{
\begin{tabular}{@{}cccccc@{}}
\toprule
\multicolumn{2}{c}{\textbf{Layer}} & \textbf{Filters} & \textbf{Size} & \textbf{Input} & \textbf{Output} \\ \midrule
        $0$ & conv & $64$ & $3 \times 3 / 1$ & $160 \times 40 \times 1$ & $160 \times 40 \times 64$ \\
        $1$ & max & & $2 \times 2 / 2$ & $160 \times 40 \times 64$ & $80 \times 20 \times 64$ \\
        $2$ & conv & $128$ & $3 \times 3 / 1$ & $80 \times 20 \times 64$ & $80 \times 20 \times 128$ \\
        $3$ & max & & $2 \times 2 / 2$ & $80 \times 20 \times 128$ & $40 \times 10 \times 128$ \\
        $4$ & conv & $256$ & $3 \times 3 / 1$ & $40 \times 10 \times 128$ & $40 \times 10 \times 256$ \\
        $5$ & conv & $256$ & $3 \times 3 / 1$ & $40 \times 10 \times 256$ & $40 \times 10 \times 256$ \\
        $6$ & max & & $2 \times 2 / 2 \times 1$ & $40 \times 10 \times 256$ & $40 \times 5 \times 256$ \\
        $7$ & conv & $512$ & $3 \times 3 / 1$ & $40 \times 5 \times 256$ & $40 \times 5 \times 512$ \\
        $8$ & batch & & & & \\
        $9$ & conv & $512$ & $3 \times 3 / 1$ & $40 \times 5 \times 512$ & $40 \times 5 \times 512$   \\
        $10$ & batch & & & & \\
        $11$ & max & & $2 \times 2 / 2 \times 1$ & $40 \times 5 \times 512$ & $40 \times 2 \times 512$ \\
         $12$ & conv & $512$ & $2 \times 2 / 2 \times 1$ & $40 \times 2 \times 512$ & \REV{$40 \times 1 \times 512$} \\[3pt]
        \toprule
    \multicolumn{2}{c}{\textbf{Layer}} & \multicolumn{2}{c}{\textbf{Input}} & \textbf{Hidden Layer} & \textbf{Output} \\ \midrule
        13 & \gls*{lstm} & \multicolumn{2}{c}{\REV{$512 \times 1 \times 40$}} & $256$ & $11$ \\ \bottomrule
    \end{tabular}}
\end{center}
\end{table}
\section{Experimental Results}
\label{sec:results}

In this section, we report the experiments carried out to verify the effectiveness of the \gls*{cnn}-based methods in the \dataset dataset and also in public datasets. 
\REV{All experiments were performed on a computer with an AMD Ryzen Threadripper $1920$X $3.5$GHz CPU, $32$ GB of RAM and an NVIDIA Titan Xp GPU ($3$,$840$ CUDA cores and $12$ GB of RAM).}

We first assess counter detection since the counter regions used for recognition are from the detection results, rather than cropped directly from the ground truth.
This is done to provide a realistic evaluation of the entire \gls*{amr} system, where well-performed counter detection is essential to achieve outstanding recognition results. 
Next, each approach for counter recognition is evaluated and a comparison between them is presented. 

Counter detection is evaluated in the \dataset dataset and also in two public datasets~\cite{vanetti2013gas,goncalves2016reconhecimento}, while counter recognition is assessed only in the \dataset dataset. This is because (i) two different sets of images were used to evaluate digit segmentation and recognition in Ref.~\citenum{vanetti2013gas}, and thus it is not possible to use these sets in the counter recognition approaches (where these stages are performed jointly); (ii) Ref.~\citenum{goncalves2016reconhecimento} performed digit recognition on a subset of their dataset which was not made publicly available.

We will finally evaluate the entire \gls*{amr} pipeline in a subset of $100$ images ($640\times480$) taken from the public dataset introduced by Vanetti et al.~\cite{vanetti2013gas}. This subset, called Meter-Integration, was used to perform an overall evaluation of the \gls*{amr} methods proposed in Refs.~\citenum{vanetti2013gas,gallo2015robust}. It should be noted that other subsets of the dataset, containing different images, were used to evaluate each \gls*{amr} stage independently and the training images (in the overall evaluation) are from these subsets~\cite{vanetti2013gas}. 
Aiming at a fair comparison, we employ the same protocol.

\subsection{Counter Detection}
\label{subsec:results_detection}

For evaluating counter detection, we employ the bounding box evaluation defined in the PASCAL VOC Challenge~\cite{everingham2010pascalvoc}, where the predicted bounding box is considered to be correct if its \gls*{iou} with the ground truth is greater than $50$\% (\gls*{iou}~$>0.5$). 
This metric was also used in previous works~\cite{nodari2011multineural,vanetti2013gas}, being interesting once it penalizes both over- and under-estimated objects.

\begin{figure*}[!htb]
	\begin{center}
	\includegraphics[width=0.85\columnwidth]{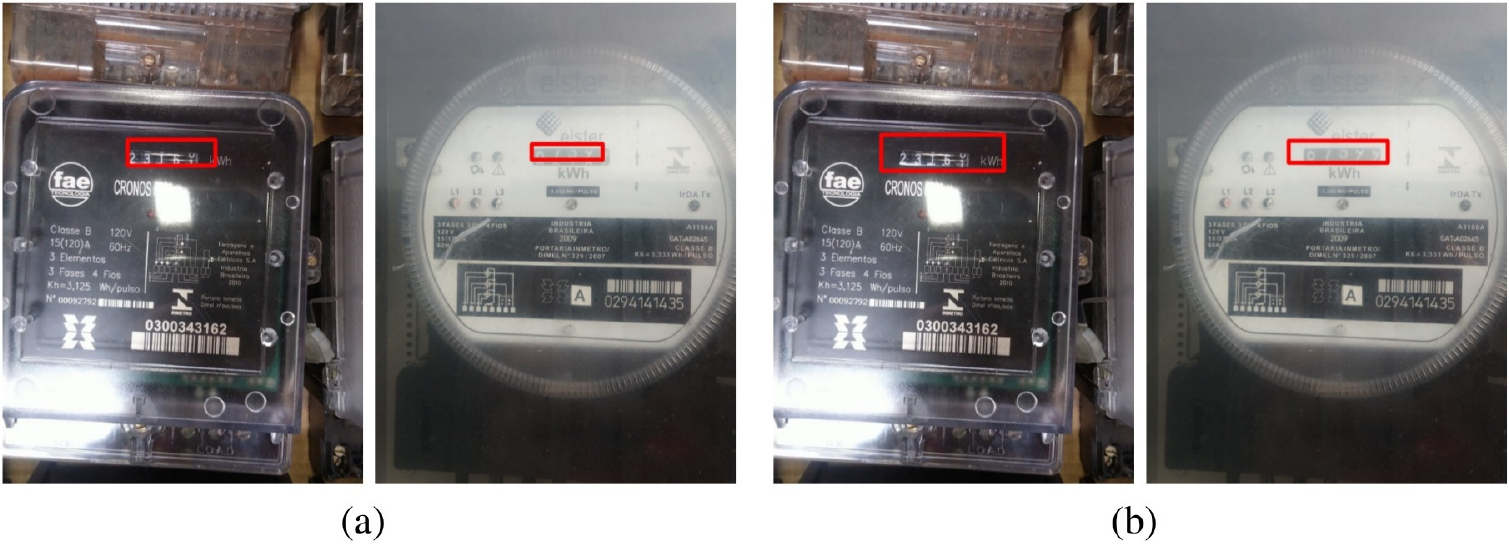}
	\end{center}
	\vspace{-3mm}
	\caption{Bounding boxes predicted by the Fast-YOLO model before~(a) and after~(b) adding the margin ($20$\%~of the bounding box~size).}
	\label{fig:detection_margin}    
\end{figure*}

According to the detection evaluation described above, the network correctly detected $99.75$\% of the counters with an average \gls*{iou} of $83\%$, failing to locate the counter in just two images~($798$/$800$). However,  in these two cases, it is still possible to recognize the digits from the detected counters, since they were actually detected (with \gls*{iou} $\le 0.5$) and all digits are within the \gls*{roi} after adding a margin (as explained in Section~\ref{subsec:proposed_detection}). In the validation set, a margin of $20$\% (of the bounding box size) is required so that all digits are within the \gls*{roi}. Thus, we applied a $20$\% margin in the test set as well. Fig.~\ref{fig:detection_margin} shows both cases where the counters were detected with \gls*{iou}~$\le$~$0.5$ before and after adding this margin. Note that, in this way, all counter digits are within the located region using Fast-YOLO. 

\captionsetup[subfigure]{position=bottom}

\begin{figure*}[!htb]
	\begin{center}
	\includegraphics[width=0.97\columnwidth]{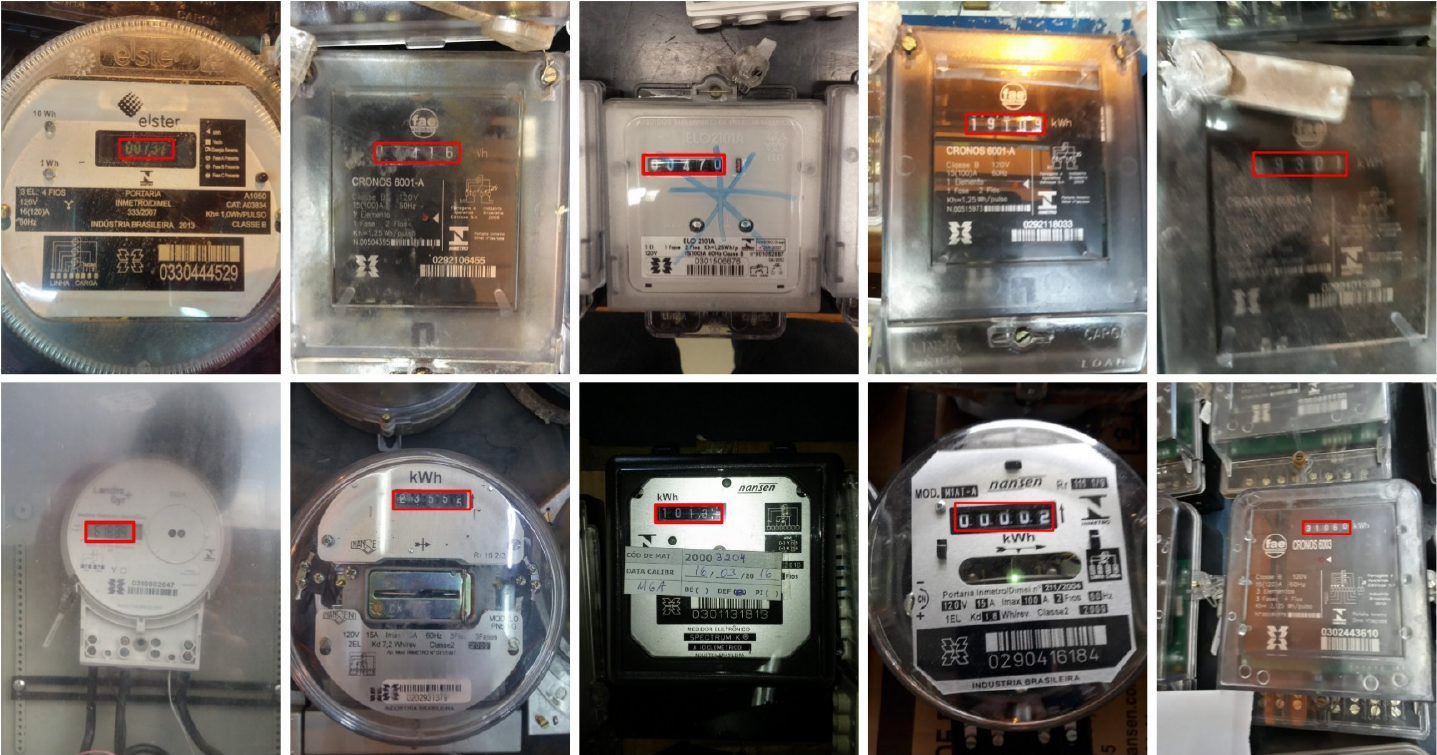}
	\end{center}
	\vspace{-2mm}
	\caption{Samples of counter detection obtained with the Fast-YOLO model in the \dataset dataset.
	}
	\label{fig:results_detection}    
\end{figure*}

Some detection results achieved by the Fast-YOLO model are shown in Fig.~\ref{fig:results_detection}. 
As can be seen, well-located predictions were attained on counters of different types and under different conditions.

In terms of computational speed, the Fast-YOLO model takes about $3.30$ ms per image ($303$ \gls*{fps}). The model was trained using the Darknet framework~\cite{darknet13} and the following parameters were used for training the network: $60k$ iterations (max batches) and learning rate~=~[$10$\textsuperscript{-$3$},~$10$\textsuperscript{-$4$},~$10$\textsuperscript{-$5$}] with steps at $25k$ and $35k$ iterations.

\subsubsection{Counter Detection on Public Datasets}

To demonstrate the robustness of Fast-YOLO for counter detection, we employ it on the public datasets found in the literature~\cite{vanetti2013gas,goncalves2016reconhecimento} and compare the results with those reported in previous works.
Vanetti et al.~\cite{vanetti2013gas} employed a subset of $153$ images of their dataset specially for the evaluation of counter detection, being $102$ for training and $51$ for testing. In Ref.~\citenum{goncalves2016reconhecimento}, a larger dataset (with $640$ images) was introduced, but no split protocol was defined.

As the dataset introduced in Ref.~\citenum{vanetti2013gas} has a split protocol, we employed the same division in our experiments. We randomly removed $20$ images from the training set and used them as validation. For the experiments performed in the dataset introduced in Ref.~\citenum{goncalves2016reconhecimento}, we perform $5$-fold cross-validation with images assigned to folds randomly in order to achieve a fair comparison. Thus, in each run, we used $384$ images~($60\%$) for training and $128$ images~($20\%$) for each validation and testing, i.e., a $3/1/1$ split protocol. 

As mentioned in the related work section, both datasets are composed of gas meter images.
Such a fact is relevant since gas meters usually have red decimal digits that should be discarded in the reading process~\cite{vanetti2013gas,gallo2015robust,goncalves2016reconhecimento,gomez2018cutting}. Therefore, we manually labeled, in each image, a bounding box containing only the significant digits for training Fast-YOLO. These annotations are also publicly available to the research community at \small \url{https://web.inf.ufpr.br/vri/databases/ufpr-amr/}\normalsize.

\begin{table}[!htb]
\caption{F-measure values obtained by Fast-YOLO and previous works in the public datasets found in the literature.}
\label{tab:detection_results}
\begin{center}
\begin{tabular}{@{}ccc@{}}
\toprule
\multirow{2}{*}{Approach} & \multicolumn{2}{c}{F-measure} \\ 
& Dataset~\cite{goncalves2016reconhecimento} & Dataset~\cite{vanetti2013gas} \\ \midrule
Nodari \& Gallo~\cite{nodari2011multineural} & $-$ & $70.00$\% \\
Gonçalves~\cite{goncalves2016reconhecimento} & $96.09$\% & $88.24$\% \\
Vanetti et al.~\cite{vanetti2013gas} & $-$ & $96.00$\% \\
\textbf{Fast-YOLO} & $\textbf{100.00}$\% & $\textbf{100.00}$\% \\[3pt] \hdashline \\[-10pt]
\REV{Fast-YOLO (\gls*{iou} $>$ $0.7$)} & \REV{$98.59$\%} & \REV{$92.16$\%} \\
\bottomrule
\end{tabular}
\end{center}
\end{table}

\begin{figure}[!htb]
	\begin{center}
	\includegraphics[width=0.91\columnwidth]{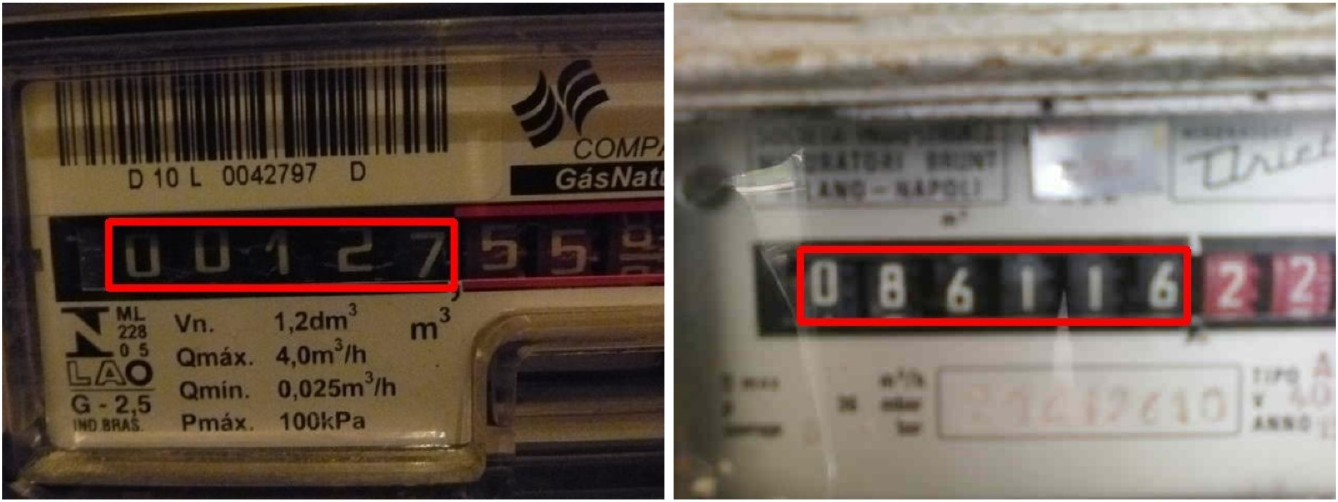}
	\end{center}
	\vspace{-2mm}
	\caption{Examples of counter detection obtained with the Fast-YOLO model. Note that the counter region in the images of the Dataset~\cite{goncalves2016reconhecimento} (left) and Dataset~\cite{vanetti2013gas} (right) is quite larger than in the \dataset dataset.} 
	\label{fig:detection_public}    
\end{figure}

The Fast-YOLO model correctly detected $100$\% of the counters in both datasets, outperforming the results obtained in previous works, as shown in Table~\ref{tab:detection_results}. 
It is noteworthy the outstanding \gls*{iou} values attained: on average $83.39  $\% in the dataset proposed in Ref.~\citenum{vanetti2013gas} and $91.28$\% in the dataset introduced in Ref.~\citenum{goncalves2016reconhecimento}. 
We believe that these excellent results are due to the fact that, in these datasets, the counter occupies a large portion of the image and the meters/counters are quite similar when comparing with the \dataset dataset. Fig.~\ref{fig:detection_public} shows a counter from each dataset detected using Fast-YOLO.

\begin{figure}
    \begin{center}
	\includegraphics[width=0.91\columnwidth]{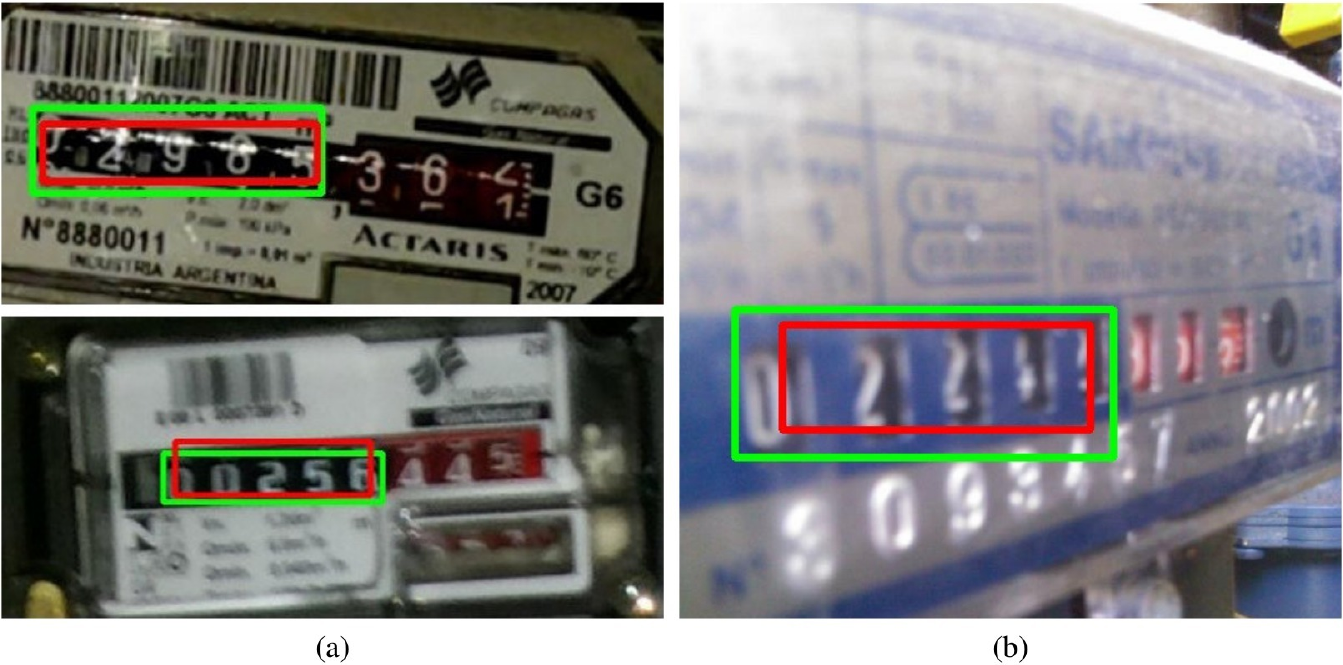}
	\end{center}	
	\vspace{-3mm} 
    \caption{Samples of counters detected with a lower \gls*{iou} with the ground truth. (a) and (b) show images of the datasets proposed in Refs.~\citenum{goncalves2016reconhecimento} and~\citenum{vanetti2013gas}, respectively. The predicted position and ground truth are outlined in red and green, respectively. Observe that all digits would be within the \gls*{roi} with the addition of a small margin.}
    \label{fig:detection_public_lower_iou}
\end{figure}

\REV{Additionally, we reported the result with a higher detection threshold (i.e., \gls*{iou}~$>$~$0.7$). It is remarkable that more than $90$\% of the counters were located with an \gls*{iou} (with the ground truth) greater than $0.7$ in both datasets. We noticed that the detections with a lower \gls*{iou} occurred mainly in cases where the meter/counter was inclined or tilted, as illustrated in Fig.~\ref{fig:detection_public_lower_iou}.
}

\subsection{Counter Recognition}
\label{subsec:results_recognition}

For this experiment, \REV{we report the mean of $10$ runs for both digit and counter recognition accuracy.} While the former is the number of correctly recognized digits divided by the number of digits in the test set, the latter is defined as the number of correctly recognized counters divided by the test set size, since each image has only a single meter/counter. 
Additionally, all \gls*{cnn} models were trained with and without data augmentation, so that we can analyze how data augmentation (described in Section~\ref{subsubsec:data_augmentation}) affects the performance of each model.

For a fair comparison, we (i) generated $300$,$000$ images and applied them for training all \glspl*{cnn} (more images were not generated due to hardware limitations); (ii) disabled the Darknet’s (hence CR-NET’s) built-in data augmentation, which creates a number of images with changed colors (hue, saturation, exposure) randomly cropped and resized; and (iii) evaluated different margin values (less than the $20$\% applied previously) in the predictions obtained by Fast-YOLO, since each approach might work better with different margin values.

The recognition rates achieved by all models are shown in Table~\ref{tab:results_recognition}. \REV{We performed statistical paired t-tests at a significance level $\alpha=0.05$, which showed that there is a significant difference in the results obtained with different models.} As expected, the results are greatly improved when taking advantage of data augmentation. The best results were achieved with the CR-NET model, which correctly recognized \REV{$94.13$\%} of the counters with data augmentation against \REV{$92.30$\%} and \REV{$87.69$\%} through \gls*{crnn} and Multi-Task Learning, respectively. This suggests that segmentation-free approaches require a lot of training data to achieve promising recognition rates, as in Ref.~\citenum{gomez2018cutting} where $177$,$758$ images were used for training.

\begin{table}[!htb]
\caption{Recognition rates obtained in the \dataset dataset using Fast-YOLO for counter detection and each of the \gls*{cnn} models for counter recognition.}
\label{tab:results_recognition}
\vspace{-2mm}
\begin{center}
\begin{tabular}{@{}ccc@{}}
\toprule
\multicolumn{1}{c}{\multirow{2}{*}{Approach}} & \multicolumn{2}{c}{Accuracy (\%)} \\
\multicolumn{1}{c}{} & Digits & Counters \\ \midrule
Multi-Task (original training set) & \REV{$24.64\pm0.25$} & \REV{$00.00\pm0.00$} \\
\gls*{crnn} (original training set) & \REV{$92.85\pm0.93$} &  \REV{$77.75\pm2.39$} \\
CR-NET (original training set) & \REV{$97.78\pm0.17$} & \REV{$91.95\pm0.52$} \\ \midrule
Multi-Task (with data augmentation) & \REV{$95.96\pm0.25$} & \REV{$87.69\pm0.40$} \\
\gls*{crnn} (with data augmentation)  & \REV{$97.87\pm0.21$} & \REV{$92.30\pm0.56$} \\
\textbf{CR-NET (with data augmentation)} & \REV{$\textbf{98.30}\pm\textbf{0.09}$} & \REV{$\textbf{94.13}\pm\textbf{0.50}$} \\ \bottomrule
\end{tabular}
\end{center}
\end{table}

It is important to highlight that it was not possible to recognize any counter when training the Multi-Task model without data augmentation. \REV{We performed several experiments reducing the size of the Multi-Task network to verify if a smaller network could learn a better discriminant function. However, better results were not achieved.}
This is because the dataset is biased and so is the recognition.
\REV{Even though the first digit has the strongest bias (given the large amount of $0$ and $1$s in that position), the other digits still have a considerable bias due to the low number of training samples. For example, the Multi-Task network may learn to predict the last digit/task as `$5$' on every occasion it sees a particular combination of the other digits that is present in the training set. In other words, the network may learn correlations between the outputs that do not exist in practice (in other applications this may be beneficial, but in this case it is not). Such a fact explains why the segmentation-free approaches had a higher performance gain with data augmentation, which balanced the training set and eliminated the undesired correlation between the outputs.}

To assess the speed/accuracy trade-off of the three \gls*{cnn} models, we list in Table~\ref{tab:results_recognition_fps} the time required for each approach to perform the recognition stage. We report the \gls*{fps} rate achieved by each approach considering only the recognition stage and also considering the detection stage (in parenthesis), which takes about $3.30$ ms per image using Fast-YOLO. 
The reported time is the average time spent processing all images, assuming that the network weights are already loaded. \MINOR{For completeness, for each network, we also list the number of parameters as well as the number of \gls*{bflop} required for a single forward pass over a single image.}

\begin{table}[!htb]
\caption{Results obtained in the \dataset dataset and the computational time required for each approach to perform counter recognition. In parentheses is shown the \gls*{fps} rate when considering the detection stage.}
\label{tab:results_recognition_fps}
\vspace{-2mm}
\begin{center}
\begin{tabular}{@{}cccccc@{}}
\toprule
Approach & \MINOR{\acrshort*{bflop}} & \MINOR{Parameters} & Time (ms) & \gls*{fps} & Accuracy (\%)\\ \midrule 
Multi-Task & \MINOR{$3.45$} & \MINOR{$209$M} & $0.6956$ & $1437$ ($250$) & \REV{$87.69\pm0.40$} \\
\gls*{crnn} & \MINOR{$2.50$} & \MINOR{$7$M} & $5.1751$ & $193$ ($118$) & \REV{$92.30\pm0.56$} \\ 
CR-NET & \MINOR{$5.37$} & \MINOR{$3$M} & $2.1071$ & $475$ ($185$) & \REV{$\textbf{94.13}\pm\textbf{0.50}$} \\ \bottomrule
\end{tabular}
\end{center}
\end{table}

The CR-NET and Multi-Task approaches achieved impressive \gls*{fps} rates. Looking at Table~\ref{tab:results_recognition_fps}, the difference between using each one of them is clear. The CR-NET model achieved an accuracy of \REV{$94.13$\%} at $475$ \gls*{fps}, while the Multi-Task model was capable of processing $1437$ \gls*{fps} with a recognition rate of \REV{$87.69$\%}. When considering the time spent in the detection stage, it is possible to process $185$ and $250$ \gls*{fps} using the CR-NET and Multi-Task models, respectively. 

\MINOR{It is worth noting that: 
(i)~even though the Multi-Task network has many more parameters than CR-NET and \gls*{crnn}, it is still the fastest one;
(ii)~the \gls*{crnn} model requires a lower number of floating-point operations for a single forward pass than the CR-NET and Multi-Task networks, however, it is still the model that takes more time to process a single image.
In this sense, we claim that there are several factors (in addition to those mentioned above) that affect the time it takes for a network to process a frame, e.g., the input size, its specific characteristics and the framework in which it is implemented. For example, two networks may require exactly the same number of floating-point operations (or have the same number of parameters) and still one be much faster than the other.}
Although much effort was made to ensure fairness in our experiments, the comparison might not be entirely fair since we used different frameworks to implement the networks and there are probably differences in implementation and optimization between them. The CR-NET model was trained using the Darknet framework~\cite{darknet13}, whereas the \gls*{crnn} and Multi-Task models were trained using PyTorch~\cite{paszke2017automatic} and Keras~\cite{chollet2015keras}, respectively.

\begin{figure}[!htb]
	\centering
	\includegraphics[width=0.86\columnwidth]{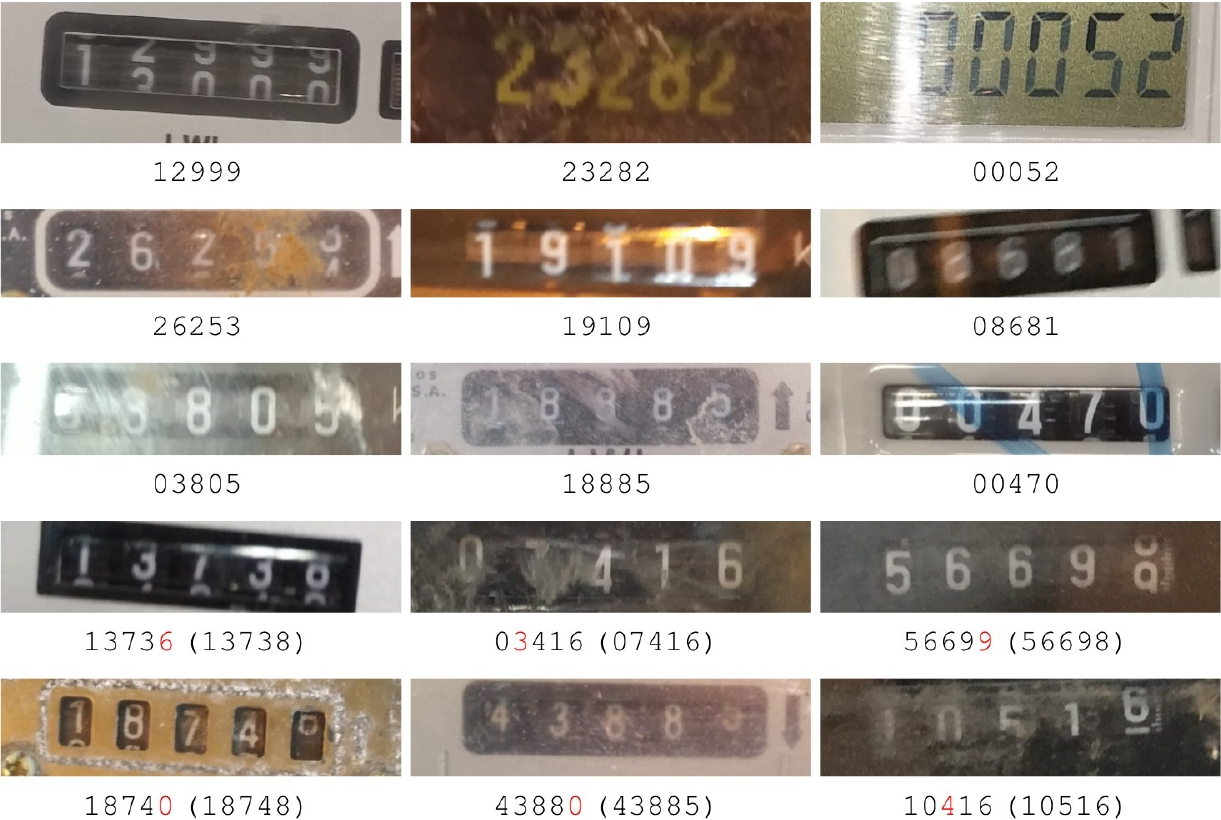}
	\vspace{1mm}
	\caption{Results obtained by the CR-NET model in the \dataset dataset. The first three rows show examples of successfully recognized counters, while the last two rows show samples of incorrectly recognized counters. Some images were slightly resized for display purposes.}
	\label{fig:results_dataset}  
	\vspace{-4mm} 
\end{figure}

Fig.~\ref{fig:results_dataset} illustrates some of the recognition results obtained in the \dataset dataset when employing the CR-NET model (i.e., the one with the best accuracy). It is noticeable that the model is able to generalize well and correctly recognize counters from meters of different types and in different conditions. Regarding the errors, we noticed that they occurred mainly due to rotating digits and artifacts in the counter region, such as reflections and dirt.

\subsection{Overall Evaluation on the Meter-Integration Subset}

The Meter-Integration subset~\cite{vanetti2013gas} was used to evaluate the \gls*{amr} methods proposed in Refs.~\citenum{vanetti2013gas,goncalves2016reconhecimento}. Thus, we decided to perform experiments on this dataset and compare the results with those obtained in both works. As previously mentioned, the training images are from other subsets of the dataset proposed in Ref.~\citenum{vanetti2013gas}. Remark that there are only $102$ and $62$ training images for counter detection and recognition, respectively.

\REV{We employ only the CR-NET model in this experiment since it outperformed both Multi-Task and \gls*{crnn} models in the \dataset dataset. The mean accuracy of $10$ runs is reported for both digit and counter recognition accuracy.} As the counters in the training set have from $4$ to $7$ digits and not a fixed number of digits, we adopted a $0.5$ confidence threshold (we report it for sake of reproducibility) to deal with a variable number of digits, instead of always considering~$5$ digits per counter.
This threshold was chosen based on $12$ validation images (i.e., $20$\%) randomly taken from the training set.
Table~\ref{tab:results_overall} shows the results obtained in previous works and using the Fast-YOLO and CR-NET networks for counter detection and recognition, respectively.

\vspace{2mm}
\begin{table}[!htb]
\caption{Results obtained in the Meter-Integration subset by previous works and using Fast-YOLO \& CR-NET.}
\label{tab:results_overall}
\vspace{-2mm}
\begin{center}
\begin{tabular}{@{}ccc@{}}
\toprule
\multicolumn{1}{c}{\multirow{2}{*}{Approach}} & \multicolumn{2}{c}{Accuracy (\%)} \\
\multicolumn{1}{c}{} & Digits & Counters \\ \midrule
Gallo et al.~\cite{gallo2015robust} (original training set) & $-$ & $85.00$ \\
Vanetti et al.~\cite{vanetti2013gas} (original training set) & $-$ & $87.00$ \\
Fast-YOLO \& CR-NET (original training set) & \REV{$97.94\pm0.85$} & \REV{$94.50\pm1.72$} \\ 
\textbf{Fast-YOLO \& CR-NET (data augmentation)} & \REV{$\textbf{99.56}\pm\textbf{0.34}$} & \REV{$\textbf{97.30}\pm\textbf{1.42}$} \\ \bottomrule 
 \end{tabular}
\end{center}
\end{table}
\vspace{-5mm}

As expected, the recognition rate accomplished by our deep learning approach
was considerably better than those obtained in previous works ($87$\% $\rightarrow \REV{94.50\%}$), which employed methods based on conventional image processing with handcrafted features. It is noteworthy the ability of both Fast-YOLO and CR-NET models to generalize with very few training images in each stage, i.e., $102$ for counter detection and $62$ for counter recognition. 
 
\begin{figure}[!htb]
	\begin{center}
	\includegraphics[width=0.81\columnwidth]{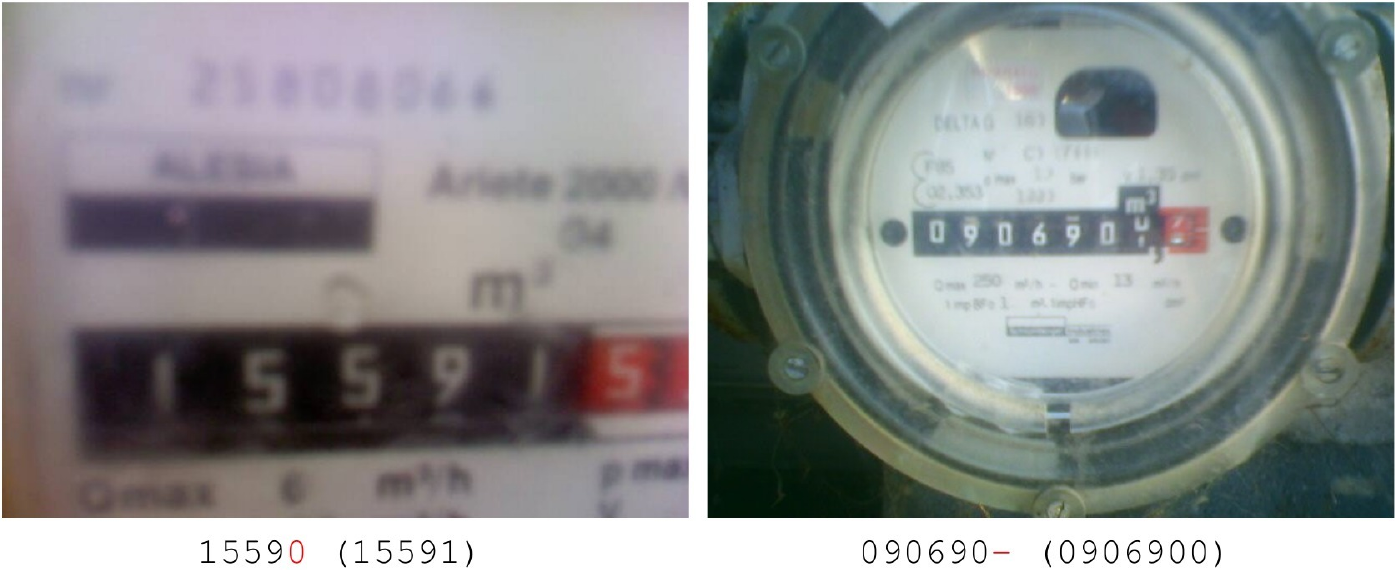}
    \end{center}
    \vspace{-1mm}
    \caption{Incorrect readings obtained with the Fast-YOLO \& CR-NET approach, where the last digit was incorrectly classified (left), and the last digit was detected/recognized with a confidence value below the threshold (right).}
	\label{fig:result_meter_integration}  
\end{figure}

The results were improved when using data augmentation, as in the experiments carried out on the \dataset dataset.
The accuracy achieved was \REV{$97.30$\%}, significantly outperforming the baselines.
\REV{It is worth noting that, on average, only $2$-$3$ counters} were incorrectly classified and \REV{generally} the error occurred in the rightmost digit of the counter. 
Two samples of errors are shown in Fig.~\ref{fig:result_meter_integration}: the last digit $1$ was incorrectly labeled as $0$ in one of the cases, probably due to some noise in the image, while in the other case the last digit was detected/recognized with confidence lower than $0.5$, apparently due to the m$^3$ text touching the digit (there were no similar examples in the training~set). 

\section{Conclusions}
\label{sec:conclusion}

In this paper, we presented a two-stage \gls*{amr} approach with \glspl*{cnn} being employed for both counter detection and recognition. The Fast-YOLO~\cite{redmon2016yolo} model was employed for counter detection, while three CNN-based approaches (CR-NET~\cite{montazzolli2017}, Multi-Task Learning~\cite{goncalves2018realtime} and \gls*{crnn}~\cite{shi2017endtoend}) were employed for counter recognition. In addition, we proposed the use of data augmentation for training the \gls*{cnn} models for counter recognition, in order to construct a balanced training set with many more examples.

We also introduced a public dataset that includes $2$,$000$ images (with $10$,$000$ manually labeled digits) from electric meters of different types and in different conditions, i.e., the \dataset dataset.
It is three times larger than the largest dataset found in the literature for this task and contains a well-defined evaluation protocol, allowing a fair comparison of different methods.
Furthermore, we labeled the region containing the significant digits in two public datasets~\cite{vanetti2013gas,goncalves2016reconhecimento} and these annotations are also publicly available to the research community.

The counter detection stage was successfully tackled using the Fast-YOLO model, which was able to detect the region containing the significant digits in all images of every dataset evaluated in this work. For counter recognition, the CR-NET model yielded the best recognition results in the \dataset dataset (i.e., \REV{$94.13$\%}), outperforming both Multi-Task and \gls*{crnn} models which achieved \REV{$87.69$\%} and \REV{$92.30$\%}, respectively. These results were attained by taking advantage of data augmentation, which was essential to accomplishing promising results. In a public dataset~\cite{vanetti2013gas}, outstanding results (i.e., an overall accuracy of \REV{$97.30$\%}) were achieved using less than $200$ images for training the Fast-YOLO and CR-NET models, significantly outperforming both baselines.

The CR-NET and Multi-Task models achieved impressive \gls*{fps} rates on a high-end graphic card. When considering the time spent in the detection stage, it is possible to process $185$ and $250$ \gls*{fps} using the CR-NET and Multi-Task models, respectively. Therefore, these approaches can be employed (taking a few seconds) in low-end setups or even in some mobile phones.

As future work, we intend to create an extension of the \dataset dataset with more than $10$,$000$ images of meters of different types and under different conditions acquired by the company's employees to perform a more realistic analysis of deep learning techniques in the \gls*{amr} context.
Additionally, we plan to explore the meter’s model in the \gls*{amr} pipeline and investigate in depth the cases where the counter has rotating digits since this is one of the main causes of errors in \gls*{amr}.
\section*{Acknowledgements}
This work was supported by grants from the National Council for Scientific and Technological Development~(CNPq) (\#~428333/2016-8, \#~311053/2016-5 and \#~313423/2017-2), the Minas Gerais Research Foundation~(FAPEMIG) (APQ-00567-14 and PPM-00540-17) and the Coordination for the Improvement of Higher Education Personnel~(CAPES) (Deep-Eyes Project). We gratefully acknowledge the support of NVIDIA Corporation with the donation of the Titan Xp GPU used for this research. 
We also thank the \acrfull*{copel} for allowing one of the authors (Victor Barroso) to collect the images for the \dataset dataset.

\bibliography{bibtex}   

\begin{thebibliography}{10}

\bibitem{shu2007study}
D.~Shu, S.~Ma, and C.~Jing, ``Study of the automatic reading of watt meter
  based on image processing technology,'' in {\em IEEE Conference on Industrial
  Electronics and Applications},  2214--2217  (2007).

\bibitem{gallo2015robust}
I.~Gallo, A.~Zamberletti, and L.~Noce, ``Robust angle invariant {GAS} meter
  reading,'' in {\em International Conference on Digital Image Computing:
  Techniques and Applications},  1--7  (2015).

\bibitem{gao2018automatic}
Y.~Gao, C.~Zhao, J.~Wang, {\em et~al.}, ``Automatic watermeter digit
  recognition on mobile devices,'' in {\em Internet Multimedia Computing and
  Service},  87--95, Springer Singapore  (2018).

\bibitem{kabalci2016survey}
Y.~Kabalci, ``A survey on smart metering and smart grid communication,'' {\em
  Renewable and Sustainable Energy Reviews} {\bf 57}, 302 -- 318  (2016).

\bibitem{vanetti2013gas}
M.~Vanetti, I.~Gallo, and A.~Nodari, ``Gas meter reading from real world images
  using a multi-net system,'' {\em Pattern Recognition Letters} {\bf 34}(5),
  519--526  (2013).

\bibitem{cerman2016mobile}
M.~Cerman, G.~Shalunts, and D.~Albertini, ``A mobile recognition system for
  analog energy meter scanning,'' in {\em Advances in Visual Computing},
  247--256, Springer International Publisher  (2016).

\bibitem{quintanilha2017automatic}
D.~Quintanilha {\em et~al.}, ``Automatic consumption reading on
  electromechanical meters using {HoG} and {SVM},'' in {\em Latin American
  Conference on Networked and Electronic Media},  11--15  (2017).

\bibitem{edward2013support}
V.~C.~P. Edward, ``Support vector machine based automatic electric meter
  reading system,'' in {\em IEEE International Conference on Computational
  Intelligence and Computing Research},  1--5  (2013).

\bibitem{zhang2016automatic}
Y.~Zhang, S.~Yang, X.~Su, {\em et~al.}, ``Automatic reading of domestic
  electric meter: an intelligent device based on image processing and
  {ZigBee/Ethernet} communication,'' {\em Journal of Real-Time Image
  Processing} {\bf 12}, 133--143  (2016).

\bibitem{lecun2015deep}
Y.~LeCun, Y.~Bengio, and G.~Hinton, ``Deep learning,'' {\em Nature} {\bf
  521}(7553), 436  (2015).

\bibitem{gomez2018cutting}
L.~Gómez, M.~Rusiñol, and D.~Karatzas, ``Cutting sayre's knot: Reading scene
  text without segmentation. application to utility meters,'' in {\em 13th IAPR
  International Workshop on Document Analysis Systems (DAS)},  97--102  (2018).

\bibitem{salamon2017deep}
J.~Salamon and J.~P. Bello, ``Deep convolutional neural networks and data
  augmentation for environmental sound classification,'' {\em IEEE Signal
  Processing Letters} {\bf 24}, 279--283  (2017).

\bibitem{goncalves2016reconhecimento}
J.~C. Gon{\c{c}}alves, ``Reconhecimento de d{\'\i}gitos em imagens de medidores
  de consumo de g{\'a}s natural utilizando t{\'e}cnicas de vis{\~a}o
  computacional,'' Master's thesis, Universidade Tecnol{\'o}gica Federal do
  Paran{\'a} - UTFPR  (2016).

\bibitem{redmon2016yolo}
J.~Redmon, S.~Divvala, R.~Girshick, {\em et~al.}, ``You only look once:
  Unified, real-time object detection,'' in {\em IEEE Conference on Computer
  Vision and Pattern Recognition},  779--788  (2016).

\bibitem{montazzolli2017}
S.~Montazzolli and C.~R. Jung, ``Real-time brazilian license plate detection
  and recognition using deep convolutional neural networks,'' in {\em 30th
  Conference on Graphics, Patterns and Images (SIBGRAPI)},  55--62  (2017).

\bibitem{goncalves2018realtime}
G.~R. Gon{\c{c}}alves, M.~A. Diniz, R.~Laroca, {\em et~al.}, ``Real-time
  automatic license plate recognition through deep multi-task networks,'' in
  {\em 31th Conference on Graphics, Patterns and Images (SIBGRAPI)},  110--117
  (2018).

\bibitem{shi2017endtoend}
B.~Shi, X.~Bai, and C.~Yao, ``An end-to-end trainable neural network for
  image-based sequence recognition and its application to scene text
  recognition,'' {\em IEEE Transactions on Pattern Analysis and Machine
  Intelligence} {\bf 39}, 2298--2304  (2017).

\bibitem{du2013review}
S.~Du, M.~Ibrahim, M.~Shehata, {\em et~al.}, ``Automatic license plate
  recognition ({ALPR}): A state-of-the-art review,'' {\em Trans. on Circuits
  and Systems for Video Technology} {\bf 23}, 311--325  (2013).

\bibitem{karatzas2015icdar}
D.~Karatzas {\em et~al.}, ``{ICDAR} 2015 competition on robust reading,'' in
  {\em International Conference on Document Analysis and Recognition (ICDAR)},
  1156--1160  (2015).

\bibitem{anis2017digital}
A.~Anis, M.~Khaliluzzaman, M.~Yakub, {\em et~al.}, ``Digital electric meter
  reading recognition based on horizontal and vertical binary pattern,'' in
  {\em Int. Conference on Electrical Information and Communication Technology},
   1--6  (2017).

\bibitem{zhao2005research}
S.~Zhao, B.~Li, J.~Yuan, {\em et~al.}, ``Research on remote meter automatic
  reading based on computer vision,'' in {\em IEEE PES Transmission and
  Distribution Conference and Exposition},  1--4  (2005).

\bibitem{elrefaei2015automatic}
L.~A. Elrefaei, A.~Bajaber, S.~Natheir, {\em et~al.}, ``Automatic electricity
  meter reading based on image processing,'' in {\em IEEE Jordan Conference on
  Applied Electrical Engineering and Computing Technologies (AEECT)},  1--5
  (2015).

\bibitem{rodriguez2014hdmr}
M.~Rodriguez, G.~Berdugo, D.~Jabba, {\em et~al.}, ``{HD MR}: a new algorithm
  for number recognition in electrical meters,'' {\em Turkish Journal of Elec.
  Engineering \& Comp. Sciences} {\bf 22}, 87--96  (2014).

\bibitem{smith2007overview}
R.~Smith, ``An overview of the {Tesseract OCR Engine},'' in {\em International
  Conference on Document Analysis and Recognition},   {\bf 2}, 629--633
  (2007).

\bibitem{nodari2011multineural}
A.~Nodari and I.~Gallo, ``A multi-neural network approach to image detection
  and segmentation of gas meter counter.,'' in {\em IAPR Conference on Machine
  Vision Applications},  239--242  (2011).

\bibitem{zhao2009design}
L.~Zhao, Y.~Zhang, Q.~Bai, {\em et~al.}, ``Design and research of digital meter
  identifier based on image and wireless communication,'' in {\em International
  Conference on Industrial Mechatronics and Automation},  101--104  (2009).

\bibitem{schmidhuber2015deep}
J.~Schmidhuber, ``Deep learning in neural networks: An overview,'' {\em Neural
  Networks} {\bf 61}, 85--117  (2015).

\bibitem{copel}
Copel, ``{Energy Company Of Paraná}.''
  \url{http://www.copel.com/hpcopel/english/}.
\newblock Accessed: 2018-04-24.

\bibitem{goncalves2016benchmark}
G.~R. Gon{\c{c}}alves, S.~P.~G. da~Silva, D.~Menotti, {\em et~al.}, ``Benchmark
  for license plate character segmentation,'' {\em Journal of Electronic
  Imaging} {\bf 25}(5)  (2016).

\bibitem{laroca2018robust}
R.~Laroca, E.~Severo, L.~A. Zanlorensi, {\em et~al.}, ``A robust real-time
  automatic license plate recognition based on the yolo detector,'' in {\em
  2018 International Joint Conference on Neural Networks (IJCNN)},  1--10
  (2018).

\bibitem{wu2017squeezedet}
B.~Wu, F.~Iandola, P.~H. Jin, {\em et~al.}, ``{SqueezeDet}: Unified, small, low
  power fully convolutional neural networks for real-time object detection for
  autonomous driving,'' in {\em IEEE Conference on Computer Vision and Pattern
  Recognition Workshops (CVPRW)},  446--454  (2017).

\bibitem{tripathi2017lcdet}
S.~Tripathi, G.~Dane, B.~Kang, {\em et~al.}, ``{LCDet}: Low-complexity
  fully-convolutional neural networks for object detection in embedded
  systems,'' in {\em IEEE Conference on Computer Vision and Pattern Recognition
  Workshops},  411--420  (2017).

\bibitem{redmon2017yolo9000}
J.~Redmon and A.~Farhadi, ``{YOLO}9000: Better, faster, stronger,'' in {\em
  IEEE Conference on Computer Vision and Pattern Recognition (CVPR)},
  6517--6525  (2017).

\bibitem{everingham2010pascalvoc}
M.~Everingham, L.~Van~Gool, C.~K.~I. Williams, {\em et~al.}, ``The pascal
  visual object classes ({VOC}) challenge,'' {\em International Journal of
  Computer Vision} {\bf 88}, 303--338  (2010).

\bibitem{imagenet2009}
J.~Deng, W.~Dong, R.~Socher, {\em et~al.}, ``Image{N}et: A large-scale
  hierarchical image database,'' in {\em Conference on Computer Vision and
  Pattern Recognition},  248--255  (2009).

\bibitem{alexeyab}
AlexeyAB, ``{YOLOv2 and YOLOv3}: how to improve object detection.''

\bibitem{alexe2012measuring}
B.~Alexe, T.~Deselaers, and V.~Ferrari, ``Measuring the objectness of image
  windows,'' {\em IEEE Transactions on Pattern Analysis and Machine
  Intelligence} {\bf 34}, 2189--2202  (2012).

\bibitem{jakub2017holistic}
J.~Špaňhel, J.~Sochor, R.~Juránek, {\em et~al.}, ``Holistic recognition of
  low quality license plates by {CNN} using track annotated data,'' in {\em
  IEEE Intern. Conference on Advanced Video and Signal Based Surveillance},
  1--6  (2017).

\bibitem{gers1997learning}
F.~A. Gers, J.~Schmidhuber, and F.~Cummins, ``Learning to forget: continual
  prediction with {LSTM},'' in {\em International Conference on Artificial
  Neural Networks},   {\bf 2}, 850--855 vol.2  (1999).

\bibitem{graves2006connectionist}
A.~Graves, S.~Fern\'{a}ndez, F.~Gomez, {\em et~al.}, ``Connectionist temporal
  classification: Labelling unsegmented sequence data with recurrent neural
  networks,'' in {\em International Conference on Machine Learning (ICML)},
  369--376  (2006).

\bibitem{darknet13}
J.~Redmon, ``Darknet: Open source neural networks in {C}.''
  \url{http://pjreddie.com/darknet/}  (2013--2019).

\bibitem{paszke2017automatic}
A.~Paszke {\em et~al.}, ``Automatic differentiation in {PyTorch},''  (2017).

\bibitem{chollet2015keras}
F.~Chollet {\em et~al.}, ``Keras.'' \url{https://keras.io}  (2015).

\end{thebibliography}
\bibliographystyle{spiejour}   


\vspace{2ex}\noindent\textbf{Rayson Laroca} received his bachelor’s degree in software engineering from the State University of Ponta Grossa, Brazil. Currently, he is a master's student at the Federal University of Paraná, Brazil. His research interests include machine learning, pattern recognition and computer vision.

\vspace{2ex}\noindent\textbf{Victor Barroso} is an undergraduate student in computer science at the Federal University of Paraná, Brazil. His research interests include machine learning, computer vision, pattern recognition and its applications.

\vspace{2ex}\noindent\textbf{Matheus A. Diniz} is a master's student at the Federal University of Minas Gerais, Brazil, where he also received his bachelor's degree in computer science. His research focuses on deep learning techniques applied to computer vision and surveillance.

\vspace{2ex}\noindent\textbf{Gabriel R. Gonçalves} is a PhD student at Federal University of Minas Gerais. He received a bachelor’s degree in computer science from Federal University of Ouro Preto, Brazil and a master’s degree in computer science from the Federal University of Minas Gerais, Brazil. His research interests includes machine learning, computer vision and pattern recognition, specially applied to smart surveillance tasks. 

\vspace{2ex}\noindent\textbf{William Robson Schwartz} is an associate professor in the Department of Computer Science at the Federal University of Minas Gerais, Brazil. He received a PhD from the University of Maryland, College Park, Maryland, USA. His research interests include computer vision, smart surveillance, forensics, and biometrics, in which he authored more than 100 scientific papers and coordinated projects sponsored by several Brazilian Funding Agencies. He is also the head of the Smart Surveillance Interest Group.

\vspace{2ex}\noindent\textbf{David Menotti} is an associate professor at the Federal University of Paraná, Brazil. He received his BS and MS degrees in computer engineering and applied informatics from the Pontifical Catholic University of Paraná, Brazil, in 2001 and 2003, respectively, and his PhD degree in computer science from the Federal University of Minas Gerais, Brazil, in 2008. His research interests include machine learning, image processing, pattern recognition, computer vision, and information retrieval.



\end{spacing}
\end{document}